\documentclass{article}

\usepackage{arxiv}

\usepackage[utf8]{inputenc} 
\usepackage[T1]{fontenc}    
\usepackage{hyperref}       
\usepackage{url}            
\usepackage{booktabs}       
\usepackage{amsfonts}       
\usepackage{amsmath}
\usepackage{nicefrac}       
\usepackage{microtype}      
\usepackage{lipsum}
\usepackage{graphicx}
\usepackage{caption}
\usepackage{subcaption}
\usepackage{placeins}
\usepackage{makecell}
\usepackage{authblk}
\usepackage[square, numbers, semicolon, sort&compress]{natbib}
\setcitestyle{authoryear,open={(},close={)}}
\graphicspath{ {./images/} }

\title{A Bayesian approach to translators' reliability assessment}
\author[1,2]{Marco Miccheli}
\author[6, 7, **]{Andrej Leban}
\author[5]{Andrea Tacchella}
\author[2,4]{Andrea Zaccaria}
\author[4]{Dario Mazzilli}
\author[3, *]{Sébastien Bratières}

\affil[1]{Sapienza University of Rome, Physics Department}
\affil[2]{Istituto dei Sistemi Complessi - CNR, UOS Sapienza, Rome}
\affil[3]{Translated SRL, Rome}
\affil[4]{Centro Ricerche Enrico Fermi, Rome}
\affil[5]{Joint Research Centre, Seville}
\affil[6]{University of California, Berkeley, Department of Statistics}
\affil[7]{University of California, Berkeley, Speech and Computation Lab}
\affil[*]{To whom correspondence should be addressed. E-mail: sebastien@translated.com}
\affil[**]{Work carried out while on an internship with Translated.}

\begin{document}
\maketitle
\begin{abstract}
Translation Quality Assessment (TQA) is a process conducted by human translators and is widely used, both for estimating the performance of (increasingly used) Machine Translation, and for finding an agreement between translation providers and their customers. While translation scholars are aware of the importance of having a reliable way to conduct the TQA process, it seems that there is limited literature that tackles the issue of reliability with a quantitative approach. In this work, we consider the TQA as a complex process from the point of view of physics of complex systems and approach the reliability issue from the Bayesian paradigm. Using a dataset of translation quality evaluations (in the form of error annotations), produced entirely by the Professional Translation Service Provider Translated SRL, we compare two Bayesian models that parameterise the following features involved in the TQA process: the translation difficulty, the characteristics of the translators involved in producing the translation, and of those assessing its quality - the reviewers. We validate the models in an unsupervised setting and show that it is possible to get meaningful insights into translators even with just one review per translation; subsequently, we extract information like translators' skills and reviewers' strictness, as well as their consistency in their respective roles. Using this, we show that the reliability of reviewers cannot be taken for granted even in the case of expert translators: a translator's expertise can induce a cognitive bias when reviewing a translation produced by another translator. The most expert translators, however, are characterised by the highest level of consistency, both in translating and in assessing the translation quality. 
\end{abstract}


\section{Introduction}
Translation Quality Assessment (TQA) is a field within Translation Studies (TS) - an interdisciplinary paradigm dealing with the systematic study of the theory, description, and application of translation
\citep{Holmes1988Translated:Studies} - that focuses on the evaluation of translated text quality, both in the case of a human translator (Human Translation - HT), and when artificial intelligence is used (Machine Translation - MT). Translation is clearly a complex process, involving several linguistic and extra-linguistic variables; the latter are associated with texts, translators, languages, and the environment. This complexity, unsurprisingly,  influences the difficulty of estimating the quality of a translation's final result \citep{Castilho2018ApproachesAssessment}. 
For this reason, translation quality has been a central topic of research in the field of translation studies for decades  \citep{Nida1964TowardTranslating, Holmes1988Translated:Studies}.  One of the main issues raised by translation scholars is the lack of an objective method of estimating translation quality. Regarding this point, House stated in 1997: "evaluating the quality of a translation presupposes a theory of translation. Thus different views of translation lead to different concepts of translational quality, and hence different ways of assessing it." \citep{House1997TranslationRevisited}

Together with the complexity of TQA, the need of having a reliable way to assess translation quality has attracted the attention of many researchers. Such necessity stems from both academic research and the interests of industry, which needs a reliable way to assess the quality of the translations produced. In the case of academic research, it has to be remarked that TQA is crucial for the evaluation of MT systems.  In a world where MT usage is becoming increasingly high \citep{Way2018QualityTranslation, Lagarda2015TranslatingTechniques}, the growing effort put into the improvement of MT systems has given rise to a need for reliable metrics with which one could evaluate the changes in performance and compare different MT engines. For example, assessing translation quality with unreliable methods can lead to dangerous and erroneous claims about actual MT performance \citep{Toral2018NoHumanParity}.
Despite this, it is not common practice to investigate the reliability of the methods used in the TQA process \citep{Han2020TranslationReview} and the actual skills of the translators involved. With this work, we therefore aim to provide a method to trace translators' typical behaviour, hopefully leading to a deeper understanding of the TQA process that assessed the quality of a translation.

From the point of view of the translation industry, it is clear that Professional Translation Service Providers - the companies that provide professional linguistic translations - need to evaluate the translators engaged. Additionally, they have to communicate with transparency the quality of the service provided to their customers in order to be competitive \citep{Martinez2014TQAmetrics}. These needs cannot be fulfilled if the reliability of the TQA process is not validated.

In the task of evaluating MT outputs, the common practice is to use human translators in making the evaluations. Such human assessment is seldom verified \citep{Castilho2018ApproachesAssessment}, especially since the linguists\footnote{In this article, we use the term "linguist" to encompass professional translators acting as both the first-pass translator of a document, and the (usually more senior) reviewer, also a professional translator.} engaged are usually expert translators. 

The intrinsic subjectivity of translation raises the main concern about the quality of its evaluation by humans \citep{Zehnalova2013TraditionAssessment}: in the cited case, an investigation into the TQA evaluation revealed that the agreement between evaluators can be so low that a further step of reconciling the evaluations is needed in order to have a reliable assessment. This is the case of \cite{Daems2013QualityMT+PE}, who created the gold standard for TQA evaluation by asking two expert translators to make an assessment and subsequently reconciling the poor agreement (only 38\% of the annotated errors were shared between the two evaluations) by letting the translators examine each other's annotations. To a physicist, such subjectivity naturally brings forth the concept of variance. We can, therefore, ask ourselves: how many different translations would we have if we asked a group of linguists (even experts) to independently translate a sentence (or to evaluate a translated sentence)? Assuming that they could not remember the previous time they worked on it, how many times would the same translator produce the exact same translation in repeat translations of a sentence? 

Despite the intrinsic subjectivity of human translation, we can strive to increase the reliability of the quality assessment by maximising the agreement between different evaluators, since this can be seen as a measure of objectivity of the evaluation.
One step towards greater objectivity \citep{YILDIZ2020ASATI,White1994TheApproaches} is to identify and minimise the sources of variability in a translation's evaluation process. If we knew, for instance, that an evaluator is usually strict in assessing a translation's quality, we could adjust their assessment in order to remove the subjective component. The recognition for the necessity of subjectivity "removal" \citep{Han2020TranslationReview,Rivera2021MTqualityReview,Turchi2014CopingWithSubjectivity} has opened a research area that, to the best of our knowledge, is still quite young.

Given the complexity of the TQA task, there have been few attempts to overcome this subjectivity: \cite{Turchi2014CopingWithSubjectivity}, in the task of assessing the quality of MT outputs, replaced human TQA with an automated system that evaluated the quality of a translated sentence by looking directly at the similarity between the MT output and its human-edited version, discarding the human error annotations. This approach of using automated metrics is useful because of the evident gain in terms of time and money spent. However, it still does not represent a valid alternative to human assessment because of its poor flexibility and reliability \citep{Chatzikoumi2020TQAreview}.
As we have instead seen, \cite{Daems2013QualityMT+PE} let the translators find a consensus by examining each other's annotations, overcoming their low agreement in independent evaluations. This is, of course, an interesting and reliable way to increase the objectivity of the evaluation, but can be unfeasible in practice when the evaluation is conducted by more than two translators. Additionally, it is obviously inapplicable in the case of a single translator.

Our work is an attempt to provide a reliable quality estimate through the application of complex systems methods, in particular within a Bayesian probabilistic framework.  Using this, we model translation subjectivity sources and remove their contributions from the final quality score, gaining insights into the factors that affected the result and maximising the objectivity of the evaluation. Such methods are expected to detect more complex patterns by looking simultaneously at all the agents involved in the process of quality assessment - such as the linguist who produced the translation, the reviewer who evaluated its quality, the source text of the translation, etc. - across all the available co-occurrences in the dataset.

Furthermore, we would like to propose a general framework applicable not only to Translation Quality Assessment but also to other fields, wherever the aggregation of subjective quantitative judgements is required, e.g. in medicine, where there could be a lack of consensus in physicians' diagnoses \citep{Jauniaux2018consensusPlacenta,rhee2016sepsisDiagnosis, Tacchella2018CollabHumanMachineMS}.

\section{Dataset}
\label{sec:dataset}
Our dataset is retrieved entirely from Translated SRL's\footnote{https://translated.com} database. The data are created through a TQA process conducted by expert human reviewers that produce one quality score for every translation job evaluated. The reviewer evaluates the translation on a Computer-Assisted Translation (CAT) tool (MateCat\footnote{https://www.matecat.com/}), i.e. a tool with a user interface that displays the translated document split into segments, contrasting the source text segments with the corresponding segments produced by the translator. Besides editing the translation, the reviewer can also annotate the errors they finds. The errors found can be classified into 4 categories (\textit{Linguistic}, \textit{Accuracy}, \textit{Style}, \textit{Client Guidelines}) and 4 severities (\textit{Preferential}, \textit{Repetition}, \textit{Minor}, \textit{Major}). The eventual quantitative evaluation of the entire document is the weighted sum of the errors found, normalised by the word count of the source document, with the weights given by the error severity: 1 for the minor errors, and 2 for the major ones (all the others are weighted with 0). The obtained score is then multiplied by 1000 for readability purposes. We thus obtain our quality score, Errors Per Thousand (EPT) since it is the number of errors per thousand words:
\begin{equation}
\label{eq:ept_definition}
    EPT = 1000~\frac{m+2M}{w}
\end{equation} 
where we used $m$ and $M$ respectively for the counts of the minor and major errors found across all the segments in the translated document, and $w$ represents the total number of words in the document. 

In this work, all absolute EPT values shown in plots have been rescaled by dividing them by their maximum value and languages are anonymised for data protection reasons.

\begin{figure}[!h]
    \centering
    \begin{subfigure}[t]{0.48\linewidth}
    \centering
    \includegraphics[width=\linewidth]{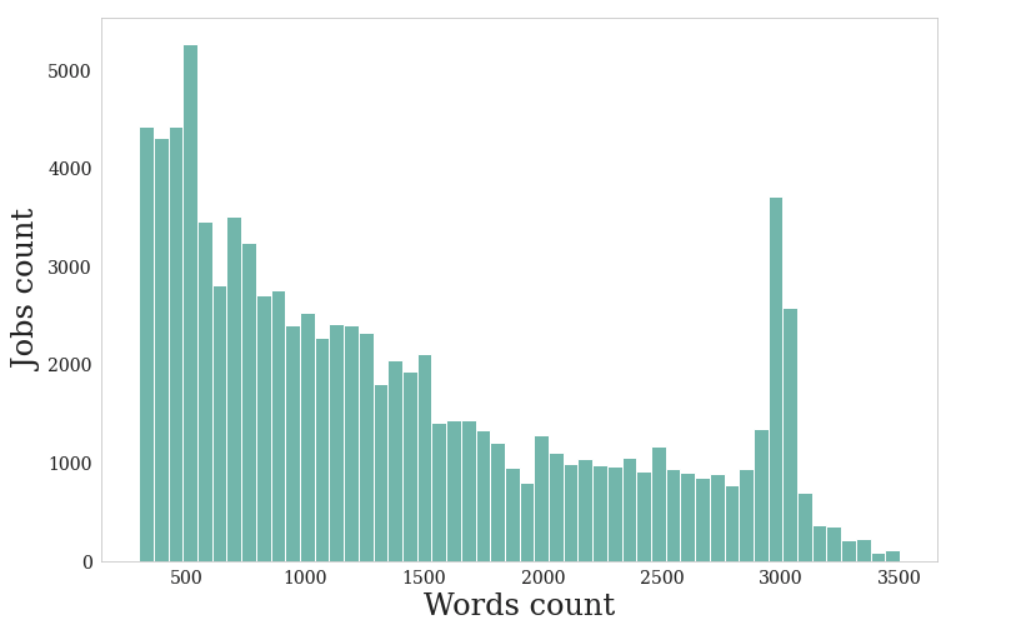}
    \caption{Histogram of the documents' distribution according to their word count.}
    \label{subfig:words_total_distribution}
    \end{subfigure}
    \hspace{2ex}%
    \begin{subfigure}[t]{0.48\linewidth}
    \centering
    \includegraphics[width=\linewidth]{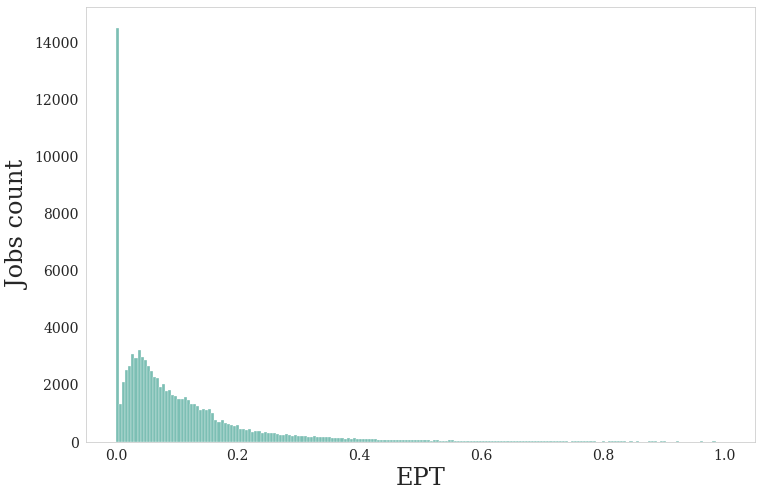}
    \caption{Histogram of the documents' distribution according to their EPT score. The high peak corresponds to the "perfect translations" (EPT = 0). EPT values have been rescaled for data protection reasons.}
    \label{subfig:ept_total_distribution}
    \end{subfigure}
    \caption{Distributions of the documents in the dataset according to the  number of words and their EPT score.}
    \label{fig:total_distributions}
\end{figure}

We have collected more than 74k reviewed translation jobs, all with US English as the source language and with 57 languages as the target, involving more than 300 translators and reviewers. All the documents have a number of words $300<w<3500$. The resulting distributions of translation jobs by number of words and by EPT score are shown in Figure \ref{fig:total_distributions}.

\FloatBarrier
\subsection{Glut of perfect translations}
\label{subsec:glut_of_zeros}
As can be noticed in Figure \ref{subfig:ept_total_distribution}, there is a relatively large fraction ($\sim 20\%$) of translations for which the reviewers did not annotate any weighted error: we refer to these cases as "perfect translations" or "zeros". This peak  seems to be detached from the rest of the distribution.

\begin{figure}
    \centering
    \begin{subfigure}[t]{0.42\linewidth}
    \centering
    \includegraphics[width=\linewidth]{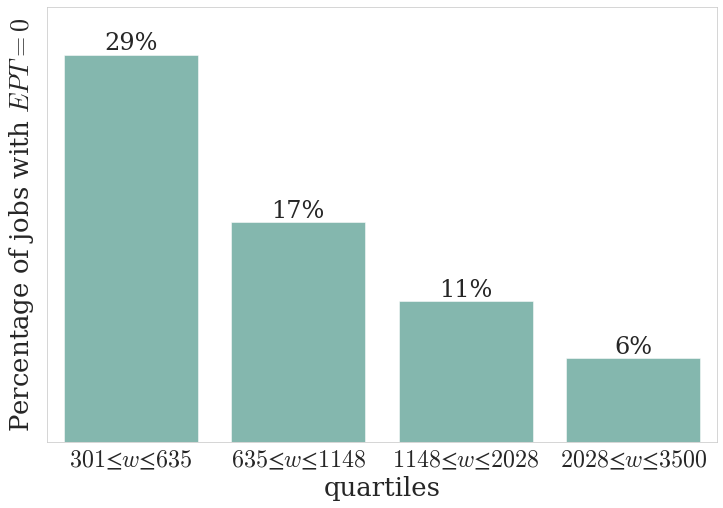}
    \caption{Percentage of translation jobs with EPT = 0 by the quartiles of the word count distribution separated by word count $w$}
    \label{subfig:glut_job_percentage_quartiles}
    \end{subfigure}
    \hspace{2ex}
    \begin{subfigure}[t]{0.45\linewidth}
    \centering
    \includegraphics[width=\linewidth]{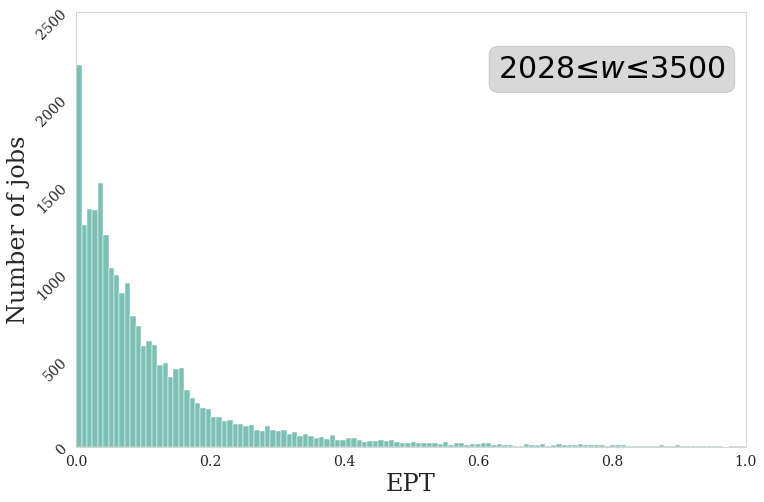}
    \caption{Distribution according to the EPT score for the last quartile of the word count distribution. EPT values have been rescaled for data protection reasons.}
    \label{subfig:glut_ept_distribution}
    \end{subfigure}
    \caption[Glut of perfect translations]{Glut of perfect translations. Even for the longest documents there is a not negligible amount of perfect translations (EPT = 0).}
    \label{fig:glut_of_perfect_translations}
\end{figure}

One could argue that this abundance of zeros can be caused solely by short documents, for which there is a smaller probability to find an error: the peak could be due to a finite-size effect in the evaluation of the translated documents. To check if the finite-size effect was the only reason for this glut of zeros, we checked the distribution of EPT for longer documents: we divided our dataset into four subgroups, using the quartiles of the distribution of documents per number of words (see fig.\ref{subfig:words_total_distribution}) as the separating criterion. However, plotting the EPT distribution for the last quartile only (documents with a number of words $2028 \leq w \leq 3500$) reveals a similar peak at EPT = 0 (see fig.\ref{subfig:glut_ept_distribution}). 

\FloatBarrier

Although the peak corresponding to the perfect translations is present also in longer documents, the percentage of zeros actually decreases with increasing document size (see Figure \ref{subfig:glut_job_percentage_quartiles}), meaning that the finite-size effect becomes less evident for longer translation jobs (and so it is true that part of this effect can be explained with the document lengths).
On the other hand, the persistent presence of the peak at $EPT=0$ also for longest documents (see Figure \ref{subfig:glut_ept_distribution}) drives us to avoid explaining the glut of zeros with just the finite-size effect, making us think about some other significant source of perfect translations. We will give our interpretations of this phenomenon in subsection \ref{subsec:hurdle}.

\section{Models}
\label{sec:models}

In our approach to model the TQA process, we consider the EPT score associated with a translation job $j$ as a measurement of the translation's latent real quality score, which we call $q_j$. We showcase two models that aim to fit the observed data by taking into account the parameters that are expected to significantly determine the outcome of the translation quality measurement, such as the intrinsic difficulty of the translation job and the translator's and reviewer's characteristics. By employing a  Bayesian probabilistic approach, we characterise translators (only with the Hurdle model, see \ref{subsec:hurdle}) and reviewers (both models) by considering them as the sources of error affecting the EPT measurements. This allows us to assess their reliability in light of an optimal set of parameters that best describes the observed EPT measurements, given our prior knowledge about the TQA process.

The optimal set of parameters is found by training the models via Markov Chain Monte Carlo (MCMC) \citep{Gelman1995BayesianDA}. With MCMC we infer the shape of the posterior $p(\boldsymbol{\Theta}|\mathcal{E})$ i.e. the probability that $\boldsymbol{\Theta}$ is the set of parameters that best fit the observed EPT data, according to our prior knowledge and the model chosen. Following Bayes' theorem, such posterior probability is proportional to the product between the likelihood and prior probabilities:
\begin{equation}
\label{eq:bayes_theorem}
    \underbrace{p(\boldsymbol{\Theta}|\mathcal{E})}_{\text{posterior}} \propto \underbrace{p(\mathcal{E}|\boldsymbol{\Theta})}_{\text{likelihood}}\underbrace{p(\boldsymbol{\Theta})}_{\text{prior}}
\end{equation}
The two models we present are encoded by the specification of the model parameters $\boldsymbol{\Theta}$, and the choice of the likelihood $p(\mathcal{E}|\boldsymbol{\Theta})$ and prior $p(\boldsymbol{\Theta})$ distributions. The models are named after the probability distributions used as the likelihood distribution.

\subsection{Gaussian model}
\label{subsec:gaussian}
The simplest Bayesian model considers the EPT scores as a random variable obtained by the measurement of the real latent quality score $q_j$. In this model, the errors are introduced solely by the reviewer who evaluates the translation. 
We have taken into account two kinds of stochastic errors introduced by the reviewer:
\begin{itemize}
    \item $\beta_r$: Offset - a systematic additive error that shifts the distribution of the EPT with respect to the real job quality $q_j$. Its interpretation is the level of \textit{strictness} of the reviewer: the higher the estimated $\beta_r$, the stricter the reviewer, giving EPT scores that are higher (i.e. the translations supposedly contain more errors) than the real quality.
    \item $\sigma_r^2$: Variance - the variability introduced by the reviewer into the values of EPT observed. For this reason, we consider $\sigma_r$ to be the reviewer's \textit{consistency}.
\end{itemize}


Our model for the posterior distribution of EPT is thus: 
\begin{equation}
    p(e_{j}|\beta_r, \sigma_r) = \mathcal{N}(q_j+\beta_r, \sigma_r^2)
\end{equation}

Since the linguists under consideration are all experts, we give the bias $\beta_r$ a zero-centred Normal prior, which is additionally weakly informative ($\sigma_{\beta}=10$). For $\sigma_r$, we set as the prior a half-Normal distribution with $\sigma_{\sigma}=1$.
The scalar $q_j$ is modelled by a Gamma distribution $\Gamma(k, \theta)$ with $k=1$ and $\theta=3$, since we wanted to have a considerable tail reflecting potentially high values of the real quality score (i.e. bad translations).

\begin{figure}[htbp]
    \centering
    \includegraphics[width=.4\linewidth]{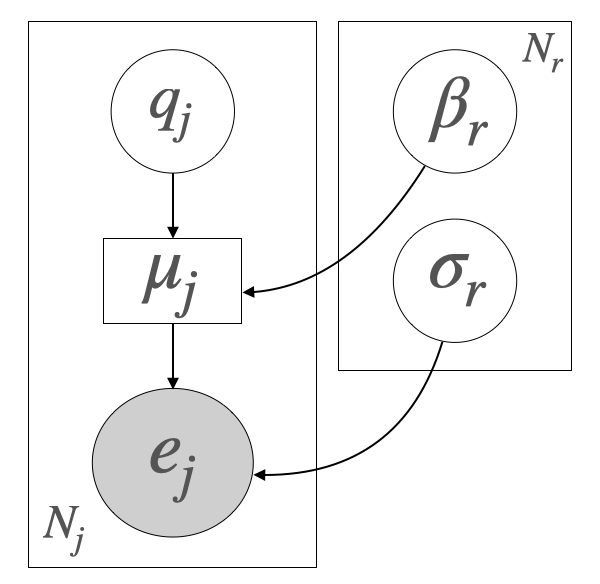}
    \caption[Gaussian model - Bayesian network]{Gaussian model Bayesian network. In this plate diagram the circles denote random variables, the squares deterministic variables, and a grey background indicates an observed random variable. Arrows indicate the influence direction between the variables. The rectangles that enclose the variables indicate their multiplicity: $N_j$ is the total number of jobs and $N_r$ is the total number of reviewers.}
    \label{fig:gaussian_model}
\end{figure}

To recapitulate, our Gaussian model is as follows
\begin{equation}
    \begin{aligned}
        q_j      &\sim \Gamma\left(1, 3\right)\\
        \beta_r  &\sim \mathcal{N}\left(0, 10^2\right)\\
        \sigma_r &\sim \mathcal{\overline{N}}\left(0, 1\right)\\
        e_{j}   &\sim \mathcal{N}\left(\mu_{j} = q_j+\beta_r, \sigma_r^2\right)
    \end{aligned}
\end{equation}
where $\mathcal{\overline{N}}(0, 1)$ is the truncated standard Normal. 
This model is illustrated in Figure \ref{fig:gaussian_model} in plate-diagram notation. Notice that we have an independent prior for each reviewer and each job, so our model contains $N_j$ parameters $q_j$ and $N_r$ parameters for both of $\beta_r$ and $\sigma_r$, where $N_j$ and $N_r$ are the numbers of jobs and reviewers, respectively.
Furthermore, since we have just one review per translation job, we only have one unique random variable $e_{jr}$ for the pair $(j,r)$ of job-reviewer, that we can call just $e_j$ - this corresponds to the two rectangles in Figure \ref{fig:gaussian_model} not intersecting.

This model allows $e_{j}<0$, which is inconsistent with the definition of EPT (Eq. \ref{eq:ept_definition}). We address this shortcoming in subsection \ref{subsec:hurdle}.

The model described here is similar to the model used in \cite{Mathur2018TowardsAnnotators}, which estimates the reliability of crowdsourcing workers, asking them to assess the quality of translated sentences against some quality-controlled sentences. As a consequence, \cite{Mathur2018TowardsAnnotators} focuses on modelling quality estimates, which is something we're not focusing on in this work.

\FloatBarrier

\subsection{Hurdle model}
\label{subsec:hurdle}
The model described above has three main modelling shortcomings: (i) it permits negative EPT values (positive by definition - see Eq.\ref{eq:ept_definition}); (ii) it does not take into account the excess of zeros as described in \ref{subsec:glut_of_zeros}; (iii) it uses a symmetric likelihood (Gaussian) for the EPT values, while the data shows an asymmetric distribution with a significant tail (see Figure \ref{subfig:ept_total_distribution}).

To address these, we introduce a more realistic model, where we replaced the Gaussian distribution with a hurdle probability distribution \citep{Cragg1971Hurdle}.
The class of hurdle distributions has two separate parts, one modelling the excess of zeros, while the other one is the distribution of non-zeros. 
They are similar to the Zero-inflated distributions \citep{hilbe2014countdata}, with the difference that the latter considers two sources of zeros, the first coming from the same distribution of non-zeros (which therefore must be defined for zero values too) and the second from another source that creates the "inflation" of zeros. 
Hurdle distributions take their name from the discontinuity which the variable has to "overcome" in order to attain a non-zero value. Such a zero and non-zero mixture distribution is governed by a Bernoulli distribution with a success parameter $\pi$.

\begin{figure}[htbp]
    \centering
    \includegraphics[width=.4\linewidth]{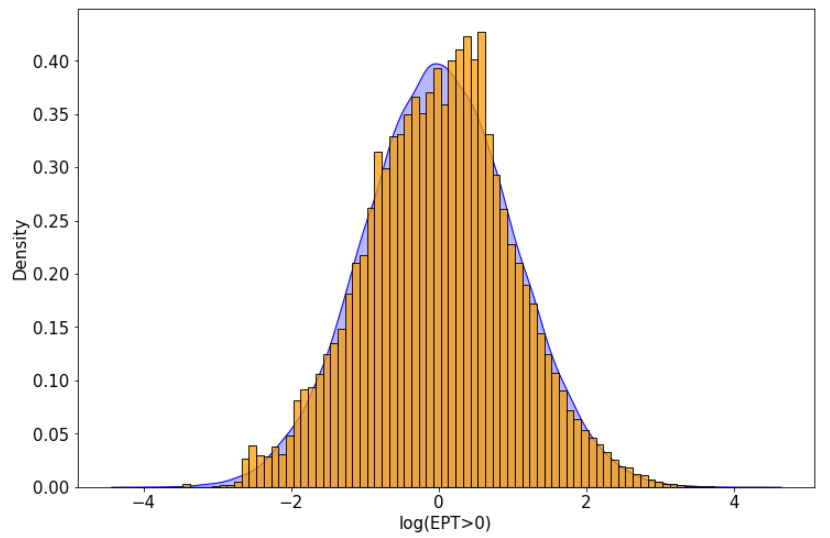}
    \caption{Gaussian fit of the standardised log EPT values. Skewness=-0.01, Kurtosis=-0.03.}
    \label{fig:Lognormal_fit}
\end{figure}

In modelling the glut of zeros described in \ref{subsec:glut_of_zeros}, we have considered the perfect translations (our zeros) as the result of a separate phenomenon that does not emerge from the regular TQA process and so, we assumed that a particular scenario must occur for the translation job to be judged as perfect. More specifically, we considered that zeros can be produced by situations in which, for example, the reviewer skips a segment out of distraction, or the translator is so skilled that they are likely to produce a perfect translation. Thus, we considered the possibility that the zeros belong to a different probability distribution than the one for the jobs with $EPT>0$: the model decides how likely it is for a specific combination of job-translator-reviewer to produce a zero or a positive value, and in the last case the EPT value is modelled by a Lognormal distribution.


In addition to the reviewer latent variables $\beta_r, \sigma_r$  considered in the Gaussian model, we also introduce the \textit{translator error propensity} $\epsilon_t$, which represents the contribution of the translator to the final EPT score, and a \textit{difficulty parameter} $d_l$ specific to the language direction $l$ (e.g. English to German).
Here, the mechanistic model underlying our interpretation of contributions is
\begin{equation}
    e_{j} = d_l \epsilon_t \beta_r
\end{equation}

In this model, therefore, the real translation job's quality, which we called $q_j$ for the Gaussian model (see \ref{subsec:gaussian}) corresponds to two multiplicative contributions of language direction and translator's skill: $q_j=d_l \epsilon_t$.
We model each of $d_l, \epsilon_t, \beta_r$ as a hurdle Lognormal. This is suggested by two factors:
(1) the shape of the observed EPT distribution for non-zero values, which can be reasonably fit by a Lognormal distribution as seen in Figure \ref{fig:Lognormal_fit} where we compare a standard normal distribution with the standardised values of $log(EPT)$, (2) the error introduced by the reviewer is better modelled as multiplicative, and not additive as stated for the Gaussian model. Therefore, if we use $\theta$ in place of any of the three random variables $d_l, \epsilon_t, \beta_r$, 

\begin{equation}
\label{eq:hurdle_function_def}
    p(\theta) = \mathcal{HL}(\pi, \mu, \sigma^2) =   
    \begin{cases}
    \pi & \theta=0\\
    (1-\pi)\mathcal{L}(\mu, \sigma^2) & \theta>0
    \end{cases}
\end{equation}
where $\mathcal{L}$ is the Lognormal distribution.

Throughout this article, random variables are indexed by $j,l,r,t$ (these stand for \textit{job}, \textit{language}, \textit{reviewer}, and \textit{translator}, respectively) and have the corresponding dimensionalities. \textbf{We emphasise that our model needs only a single observation for each triplet $(j,r,t)$ to condition random variables attached to various $t,r$.} Moreover, we train one model for every language direction $l$ separately.

Using the facts that (i) products of independent Bernoulli random variables are Bernoulli, (ii) products of independent log-normal random variables are log-normal and (iii) there is no reason to think at the parameters as mutually dependent, we can \textbf{collapse} the model and write
\begin{equation}
\label{eq:hurdle_likelihood_combined}
    \begin{cases}
        p(e_{j}) = \mathcal{HL}(\pi_{j}, \mu_{j}, \sigma_{j}^2)\\
        \pi_{j} = 1-(1-\pi_l)(1-\pi_t)(1-\pi_r)\\
        \mu_{j} = \mu_l + \mu_t + \mu_r\\
        \sigma_{j}^2 = \sigma_l^2+\sigma_t^2+\sigma_r^2
    \end{cases}
\end{equation}
where $\mu_l$ is identical for jobs from the same language.

We now give suitable priors to all ${l,t,r}$-related variables. To avoid too heavy tails, we give $\sigma_l, \sigma_t, \sigma_r$ informative priors: a truncated normal distribution $\mathcal{\bar{N}}\left(0.5, 0.25\right)$.
We give $\pi$ beta-distributed priors to constrain $\pi$ on the interval $[0,1]$. We choose $\mathcal{B}(2, 5)$ for $\pi_l$ and $\pi_t$ and $\mathcal{B}(1.5, 5)$ for $\pi_r$, to give the latter a lower mode, expecting that a perfect translation should be primarily due to the job difficulty (in the language direction $l$) or the translator's skill, rather than due to the reviewer's negligence. In the case that the total absence of errors in the evaluation process is mostly attributable to the reviewer, the most likely cause would be that they did not actually work on the quality assessment.

\begin{figure}
    \centering
    \includegraphics[width=.3\linewidth]{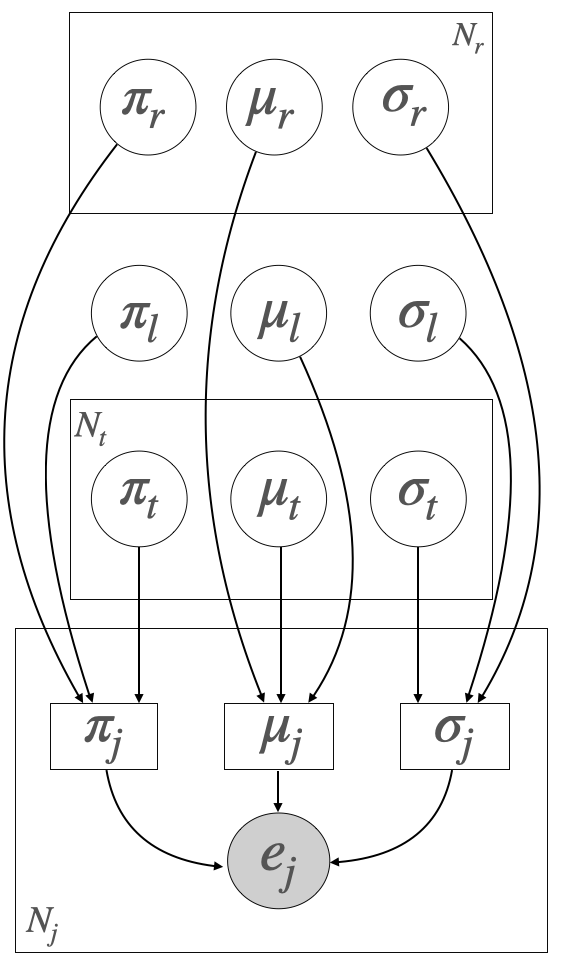}
    \caption[Hurdle model - bayesian network]{Hurdle model Bayesian network: circles enclose random variables and squares the deterministic ones. The arrows indicate the direction of the influence between variables and the grey-filled circle indicates the observed variable. Here $N_j$ stands for the total number of jobs, $N_t$ for the total number of translators, and $N_r$ for the total number of reviewers. The rectangles enclose the variables with the same multiplicity. The reader can imagine the entire figure as contained in a plate indexed by $l$ since the models are independent for each language.} 
    \label{fig:hurdle_model}
\end{figure}

We end up with the following probability model:

\begin{equation}
\centering
    \begin{aligned}
       &\pi_{l}      \sim \mathcal{B}\left(1.5, 5\right)\\
       &\pi_{t}      \sim \mathcal{B}\left(1.5, 5\right)\\
       &\pi_r          \sim \mathcal{B}\left(2, 5\right)\\
       &\mu_{v \in \{l,t,r\}}    \sim \mathcal{N}\left(0, 1\right)\\
       &\sigma_{v \in \{l,t,r\}} \sim \mathcal{\overline{N}}\left(0.5, 0.25\right)
    \end{aligned}
    \qquad\quad
    \begin{aligned}
       &\pi_{j}          = 1-(1-\pi_l)(1-\pi_t)(1-\pi_r)\\
       &\mu_{j}          = \mu_l + \mu_t + \mu_r\\
       &\sigma_{j}^2     = \sigma_l^2+\sigma_t^2+\sigma_r^2\\
       &e_{j}         \sim \mathcal{HL}\left(\pi_{j},\mu_{j}, \sigma_{j}^2\right)
    \end{aligned}
\end{equation}
as illustrated in a plate diagram in Figure \ref{fig:hurdle_model}. 

We conclude the illustration of the Hurdle model by giving our interpretations of the parameters in view of the next section:
\begin{itemize}
    \item $\pi_{v \in \{l,t,r\}}$ - probability contributions to the final $\pi_j$, the probability of observing a perfect translation. $\pi_l$ here is the overall difficulty of translation in the language pair $l$, while $\pi_t$ and $\pi_r$ are the analogous probabilities for the individual translator and reviewer, respectively. The higher the $\pi$, the more likely the outcome of the TQA process will be a perfect translation (EPT = 0).
    \item $\mu_{v \in \{l,t,r\}}$ - contributions to the mean $\mu_j$ of the underlying normal distribution of $log(e_j)$. The higher $\mu_t$, the worse the skill of the translator it represents; the higher the $\mu_r$, the stricter the reviewer (if $\mu_r = 0$ the reviewer is unbiased); the higher the $\mu_l$, the more difficult it is to produce a high-quality translation in the language pair $l$.
    \item $\sigma_{v \in \{l,t,r\}}$ - contributions to the standard deviation $\sigma_j$ of the underlying normal distribution of $log(e_j)$. The higher the $\sigma_t$, the more inconsistent in their effect on the quality of the translations the translator $t$ is. 
    The higher the $\sigma_r$, the noisier the in their effect on the EPT the reviewer is, regardless of the translator's skill who produces the translation and the difficulty of the job (we will refer to the reviewers with high variance as "inconsistent").
\end{itemize}
\FloatBarrier


\section{Validation}
Our final goal is to use the inferred parameters of the models to extract reliable insights into the features affecting the TQA process: the job's difficulty, the translator's skill, and the reviewer's behaviour. Before interpreting the inferred parameters and characterising translators and reviewers, we must check the reliability of the inference process in order to understand whether our modelling can reasonably reproduce realistic situations. 
We cannot directly measure how close our inferred parameters are to the true values of the quantities they represent, since there is no actual external (to the model) evidence of that. This measurable evidence that we're lacking is called \textit{ground truth} in the Machine Learning domain, and the modelling setting in which the ground truth is missing is called \textit{unsupervised} since it is not possible to calibrate the model by comparing predictions to observed outcomes.

In order to address the challenge posed by the lack of ground truth, we built a framework in which we can \textit{indirectly} compare our results with an available external source of information. The reason behind the term \textit{indirectly} lies in the idea that we can use the values of the latent parameters (for which we have no direct evidence)inferred by the models to generate data for which we have a ground truth to compare it with. 
For example, we are not aware whether a reviewer actually has a strict and consistent behaviour, but, if our model can synthesise new EPT data by sampling from the inferred posterior that reasonably approximate the reviewer's observed behaviour, then we can be confident that we are correctly modelling the data with the right parameters.
The better the model outcomes approximate the truth, the more meaningful the inferred parameters, and so the better will be the performance of the model when using it to characterise properties of the linguists and texts. 

Influenced by this idea, we validate our results with two experiments:
\begin{enumerate}
\item \textbf{Posterior predictive check} - once we have trained the model and inferred the posterior for all the parameters, we measure how well the models reproduce the raw EPT data observed in the training phase by drawing samples from the inferred posterior probability distributions.
\item \textbf{Skills retrieval} - we take advantage of external information about translators' skills: we are given information on which translators are certified as skilled, so we can check the ability of the model to assign a higher level of skill to the translators who have certified expertise;
\end{enumerate}

\subsection{Posterior predictive check}
Based on the observation that "the observed data should look plausible under the posterior predictive distribution" \citep{Gelman1995BayesianDA}, we draw new EPT samples  $\mathcal{E}^{rep}$ from the \textit{posterior predictive distribution}
\begin{equation}
\label{eq:posterior_predictive}
p(\mathcal{E}^{rep}|\mathcal{E}) = \int p(\mathcal{E}^{rep}|\mathbf{\Theta})\,p(\mathbf{\Theta}|\mathcal{E})\,d\mathbf{\Theta}
\end{equation}
where $p(\mathcal{E}^{rep}|\mathbf{\Theta})$ is the model likelihood, while  $p(\mathbf{\Theta}|\mathcal{E})$ is the posterior. 
The integral is not usually calculated analytically, but instead approximated through a Monte Carlo estimate based on samples from the posterior. 

In order to compare simulated against observed data, we have taken the following features of the distribution as a benchmark:
\begin{itemize}
\item number of zeros - as we pointed out in \ref{subsec:glut_of_zeros}, a distinctive feature of our dataset is the one we called the "glut of zeros". Since we are interested in the interpretation of the source of this phenomenon, we want to be sure that the model is able to reproduce such inflation of jobs with EPT = 0;
\item Kullback-Leibler divergence (KL) - we compare the overall shape of the generated distribution with the shape of the observed EPT through KL.
\end{itemize}

\subsubsection{Results}
\begin{figure}[!h]
    \centering
    \includegraphics[width=.8\linewidth]{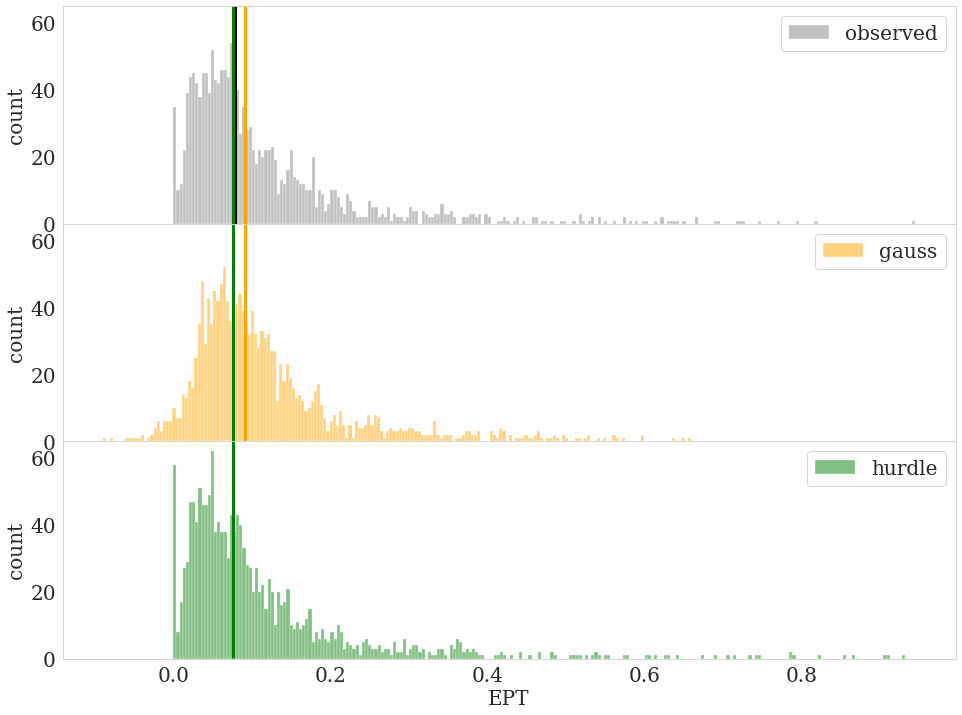}
    \caption[Posterior predictive check: example of distributions]{Posterior predictive check: examples of replicated distributions for a single language-pair, created with Gaussian (orange) and Hurdle (green) models compared to the observed distribution (grey). The vertical line is the median of the distribution, with the respective colour. All the bins in the histogram have the same width which makes it easier to visualise the differences. EPT values have been rescaled for data protection reasons.}
    \label{fig:ppc_example_jap}
\end{figure}

First, we split the dataset by language and run the inference, obtaining per-language posteriors of the parameters for both the Gaussian and Hurdle models. Then, using the same combinations of job-translator-reviewer, we draw new samples from the posterior predictive distribution defined in Eq. \ref{eq:posterior_predictive}, i.e. we generate a simulated EPT value for each job of the language pair. For each language pair, we produce 1000 replications of the language pair subsets of EPT dataset, both with the Gaussian model and with the Hurdle one, and we obtained a set of distributions, shown in Figure \ref{fig:ppc_example_jap}. 

Upon inspection, we notice that, as suspected, the Gaussian model suffers from some modelling weaknesses: its likelihood distribution (and so its posterior predictive one) is not bounded by zero, so it allows for negative EPT, which is wrong by definition (see Eq. \ref{eq:ept_definition}), and it actually exhibits this behaviour. In spite of this, for some generated data, the Gaussian model shows a median (the orange vertical line in Figure \ref{fig:ppc_example_jap}) above that of the observed data. 

On the other hand, the Hurdle model seems to replicate the presence of perfect translations, producing a peak at $\text{EPT}=0$ quite similar to the one in the observed distribution. Additionally, it accounts for the presence of extremely large values of EPT in the observed data, while the Gaussian model generates essentially no observations above 0.7 relative EPT. 

We have avoided a quantitative assessment of the ratio of zeros generated in the case of the Gaussian model, since visually examining the posterior distributions alone confirms there is no sharp peak at EPT = 0. This is, of course, expected given the structure of the model (see \ref{subsec:gaussian}): it is free to produce continuous values of EPT without any constraints, and so it does not provide inflation of zeros.

\begin{figure}[!h]
    \centering
    \includegraphics[width=\linewidth]{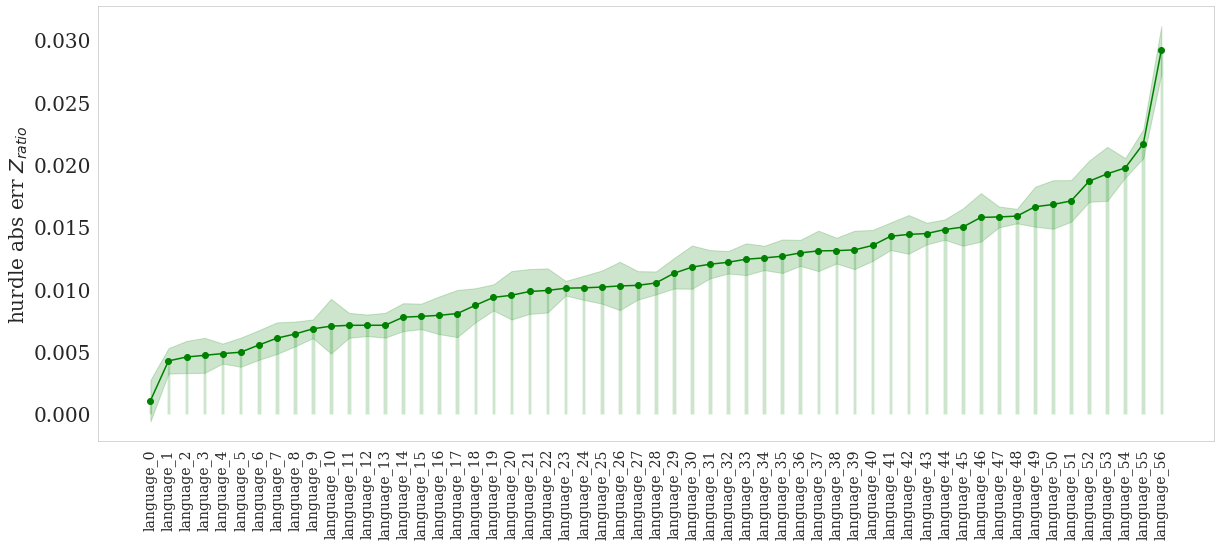}
    \caption[Posterior predictive check: ratio of zeros]{MAE of the $Z_{ratio}$, representing the deviation from the ratio of zeros in the real EPT distribution. The shaded area is the standard deviation of the absolute errors, and the languages on the x-axis are sorted by MAE, ranging from a difference of less than 0.1\% of zeros on the left to about 2.7\% on the right. Languages have been anonymised for data protection reasons.}
    \label{fig:ppc_zero_ratio}
\end{figure}

In the case of the Hurdle model, however, we were interested in how well it can fit the data by adjusting the parameter $\pi$, which is the probability that an EPT observation is exactly zero. As our metric of agreement between the generated and the observed data in this aspect we took the absolute difference between the proportion of zeros generated by the Hurdle model ($N_z^{(rep)}$) and the proportion of zeros present in the corresponding real data ($N_z^{(real)}$). Finally, we averaged the metric over the 1000 replications mentioned above, giving us the mean absolute error:
\begin{equation}
    MAE(Z_{\text{ratio}}) = \frac{1}{1000}\sum_{rep=1}^{1000}|Z_{\text{ratio}}^{(rep)}-Z_{\text{ratio}}^{(real)}|
\end{equation}
where $Z_{\text{ratio}} =\frac{N_z}{N_j}$.
In Figure \ref{fig:ppc_zero_ratio} we plot $MAE(Z_{ratio})$, language by language, and, for the sake of visualisation, we sort languages from the best-MAE language (language\_0, $\text{MAE}(Z_{ratio}) < 0.1\%$), to the worst-MAE language, language\_56, for which the model scored an absolute error in percentage of zeros of about 2.7\%.

We now compare the generated and observed data on the Kullback-Leibler divergence between their empirical distributions, calculated as $$KL(p,q) = - \sum_x p(x) \log\left(\frac{p(x)}{q(x)}\right),$$ where we plug in the observed sample for  $p(x)$ and the generated one for $q(x)$. It represents the average extra information (in nats) needed to encode $q$ assuming that the data is actually distributed as $p$.

\begin{figure}[!h]
    \centering
    \includegraphics[width=.8\linewidth]{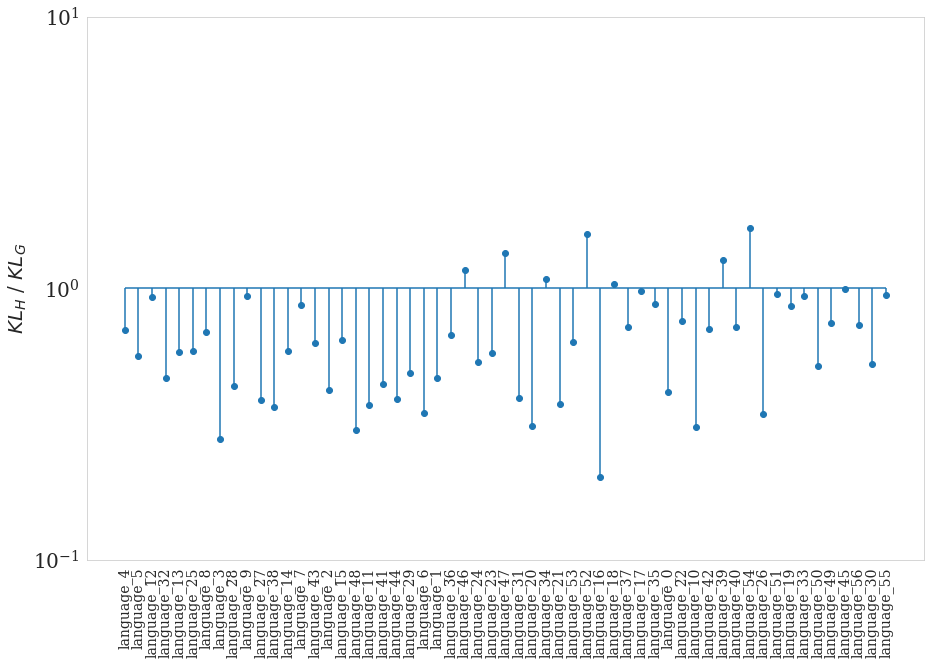}
    \caption[Posterior predictive check: KL divergence]{The ratio $\frac{KL_H}{KL_G}$ of the  Kullback-Leibler divergences between the observed distribution of EPT and the ones reproduced by the models: for most of the languages, the divergence of the Gaussian model ($KL_G$) is greater than that of the Hurdle model ($KL_H$), suggesting that for those languages, the Hurdle model fits the distribution better than the Gaussian one. The y-axis is log-scale and the horizontal line represents a "tie" between the two models.}
    \label{fig:KL_ppc}
\end{figure}

The Hurdle model exhibits lower KL than the Gaussian model, as can be observed in Figure \ref{fig:KL_ppc}, which shows the average (over the 1000 replications) ratio between the two divergences: 
$$\frac{KL_H}{KL_G}=\frac{KL(R, H)}{KL(R, G)}$$ 
with $R$, $H$, and $G$ being the empirical probability distributions (over the EPT values) of the real data, and the samples generated by the Hurdle and Gaussian models, respectively. For most languages, the ratio between $KL_H$ and $KL_G$ is lower than 1 (for which we have a tie - the horizontal line in the plot), indicating that the Hurdle model generates from a distribution that matches the observed data distribution more closely than the Gaussian model%
\footnote{In Figure \ref{fig:KL_ppc} some languages are missing for computational reasons. Sometimes it was hard to find the right binning to build a proper normalised probability distribution over the replicated EPT values provided by the models.}.%
To present a clearer visual indication of the relative deviations of the proportions of the two distances from equality (i.e., 1), we present the plot in log y-scale.

To summarise, we use the posterior predictive check to find any discrepancies in the generated data with respect to the real data and conclude that the Hurdle model reproduces the observed distribution well, matching it both in the inflation of zeros and in the shape of the rest of the distribution. On the other hand, the Gaussian model suffers from some issues, like a non-zero probability for negative EPT, as was apparent at the time of its conception. Nevertheless, the Gaussian model fits the data reasonably well, since we can observe that the discrepancies between its results and the ones produced by the Hurdle model are not that strong, even overperforming it for some languages - i.e. the ones with the KL ratio  above 1 (see Figure \ref{fig:KL_ppc}).

\FloatBarrier

\subsection{Skills retrieval}
The team of translators used by Translated is composed of expert translators, all native speakers of the target language, and specialised in marketing texts, the topic selected for the extraction of our dataset. Language teams leads are selected based on their experience by Translated's project management staff. We will denote such individuals as of skill level 1 (L1). Each is designated a deputy, of skill level 2 (L2). In our study, this selection process provides independent information about translators' skills. It can, therefore, be used to validate our models. 

\begingroup
\renewcommand{\arraystretch}{1.2}
\begin{table}[!h]
\centering
\begin{tabular}{@{} rccc @{}}    \toprule
\emph{group} & \emph{rev only} & \emph{tra only} & \emph{both}\\\midrule
skilled (L1)    & 38\% & 0  & 62\% \\ 
skilled bkp (L2) & 13\%  & 0  & 87\% \\  
unknown skill & 24\% & 12\% & 64\%\\\bottomrule 
\end{tabular}
\caption{Dataset details for the percentage of linguists per role (translator/reviewer) and skill level}
\label{tab:linguists_roles_skills}
\end{table}
\endgroup

Since most of the translators involved in the experiment act as  both translator and reviewer (cf. Table \ref{tab:linguists_roles_skills}), we can check how the above translating skill assignments reflect on the linguist characteristics when acting in the role of reviewer. In this validation, we use only the Hurdle model which takes into account the contribution of both reviewers and translators, while the Gaussian model does not. 

We split the entire population of translators and reviewers into:
\begin{itemize}
    \item Level 1: they are the most reliable translators as evaluated by Translated, and the most trusted reviewers. In the following figures, they will be shown in green on the plots.
    \item Level 2: hey are supposed to be more skilled than the average translator on a team, but sometimes they are less certified than the L1. As they act as deputies, we also refer to them as "skilled bkp" (backups). They will be represented in orange.
    \item Unknown skill: the rest of the population of translators and reviewers. They will be represented in grey.
    \begin{itemize}
        \item translator only: the translators that are in the "unknown skill" category and are not used as reviewers. Represented in light blue. 
        \item reviewer only: the reviewers that are in the "unknown skill" category and are not used as translators. Represented in pink.
    \end{itemize}
\end{itemize}

In the dataset, these linguists can be present in both roles - translator or reviewer -  or in just one of them. The breakdown of roles and skill levels is presented in Table \ref{tab:linguists_roles_skills}. On the plots, the  "Unknown skill" category alone is broken down further, with pink representing the reviewer-only category and light blue the translator-only category; both the skilled translators - L1 -  and the skilled bkp - L2 -will be represented with green and orange respectively, regardless of the role they usually assume.

As a reminder, the Hurdle model assumes that the final EPT score is composed of three multiplicative contributions: 
\begin{itemize}
    \item $d_l$ - the job difficulty in the language direction $l$, distributed as $d_l \sim \mathcal{HL}(\pi_l \mu_l, \sigma_l)$
    \item $\epsilon_t$ - the translator error propensity, distributed as $\epsilon_t \sim \mathcal{HL}(\pi_t, \mu_t, \sigma_t)$
    \item $\beta_r$ - the reviewer bias, distributed as $\beta_r \sim \mathcal{HL}(\pi_r, \mu_r, \sigma_r)$
\end{itemize}

where the distribution $\mathcal{HL(\pi, \mu, \sigma)}$ has been defined in Eq. \ref{eq:hurdle_function_def}. Given that the final EPT is a product of three random variables $d_l$, $\epsilon_t$, $\beta_r$, it is clear that both translators and reviewers can apply a contraction (or dilation) to the underlying job difficulty. Such an effect is determined by the parameters of the distribution $\epsilon_t$ for the translators and $\beta_r$ for the reviewers. We will thus use the two triplets of parameters - \{$\pi, \mu, \sigma$\} for $\epsilon_t$ and $\beta_r$ -  to characterise the properties of the linguists in the two roles. 

\subsubsection{Results}
\begin{figure}[!h]
    \centering
    \begin{subfigure}[t]{0.40\linewidth}
    \centering
    \includegraphics[width=\linewidth]{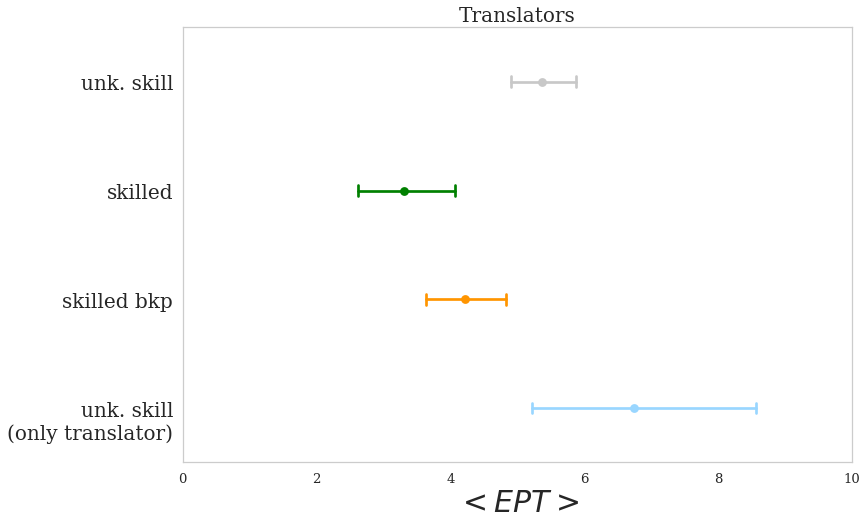}
    \caption{Translators: the averages of the mean EPT observed per translator. The higher the $<\text{EPT}>$, the lower we can assume the translator's skill to be.}
    \label{subfig:translators_features_ci_EPT}
    \end{subfigure}
    \hspace{1em}%
    \begin{subfigure}[t]{0.40\linewidth}
    \centering
    \includegraphics[width=\linewidth]{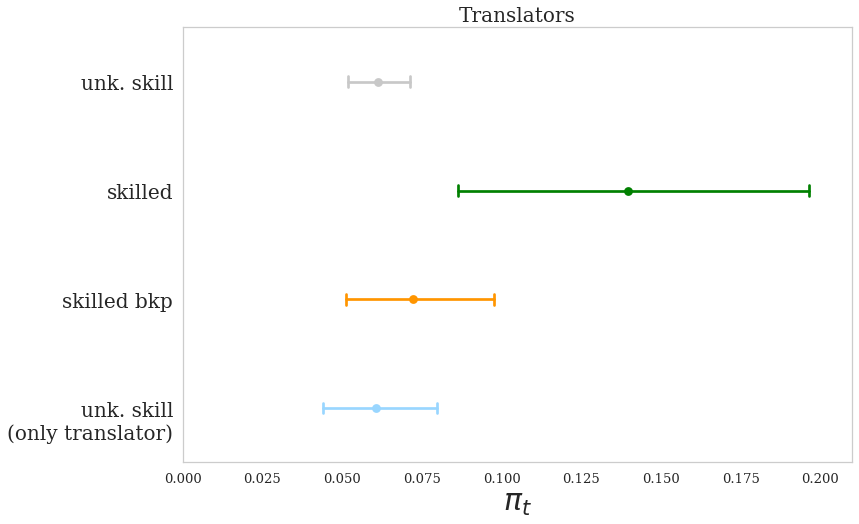}
    \caption{Translators: the averages of the mean $\pi_T$ per translator. This is the probability that the translator produces a perfect translation.}
    \label{subfig:translators_features_ci_piT}
    \end{subfigure}
    \begin{subfigure}[t]{0.40\linewidth}
    \centering
    \includegraphics[width=\linewidth]{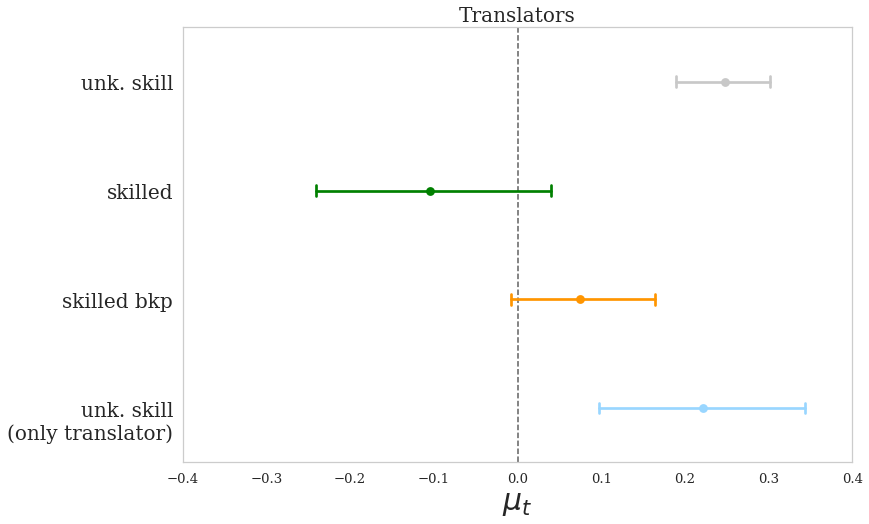}
    \caption{Translators: the averages of the mean $\mu_T$. This is a parameter of the Lognormal distribution of $\epsilon_T$ - the error propensity of the translator. The higher the $\mu_T$, the lower we interpret the translator's skill to be. The dashed vertical line at $\mu_T=0$ represents the case when the final quality score is unaffected by the translator. }
    \label{subfig:translators_features_ci_muT}
    \end{subfigure}
    \hspace{1em}
    \begin{subfigure}[t]{0.40\linewidth}
    \centering
    \includegraphics[width=\linewidth]{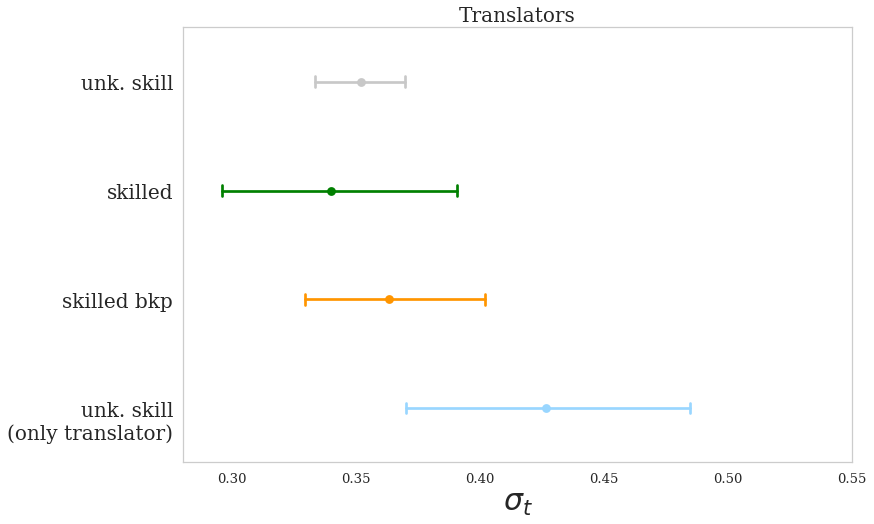}
    \caption{Translators: the averages of the mean $\sigma_T$ parameter of the Lognormal distribution of $\epsilon_T$ per translator - the error propensity of translators. The lower the $\sigma_T$, the more consistent we assume the translator to be in his propensity of introducing errors.}
    \label{subfig:translators_features_ci_sigmaT}
    \end{subfigure}
    \caption[Translators assessment - CI per skill]{Translator assessment: 95\% confidence intervals  for every parameter of the Hurdle model associated with the translators. The population of translators has been split into the language leads (green), their backups (orange), and the rest:  grey for the ones who act both as translators and reviewers at Translated, and light blue for the ones who don't appear in the dataset as reviewers.}
    \label{fig:translators_features_ci}
\end{figure}

\begin{figure}
    \centering
    \begin{subfigure}[t]{.48\linewidth}
    \includegraphics[width=\linewidth]{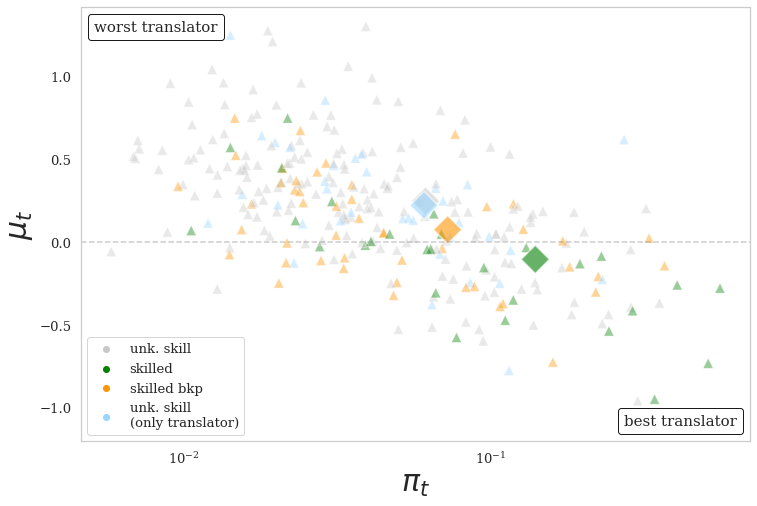}
    \caption{Translators: the higher $\pi_T$, the higher the probability of producing perfect translations. The lower $\mu_T$, the lower the probability of making errors even when the translation is not perfect.}
    \vspace{2ex}
    \label{subfig:translators_features_muT_piT}
    \end{subfigure}
    \hfill
    \begin{subfigure}[t]{.48\linewidth}
    \includegraphics[width=\linewidth]{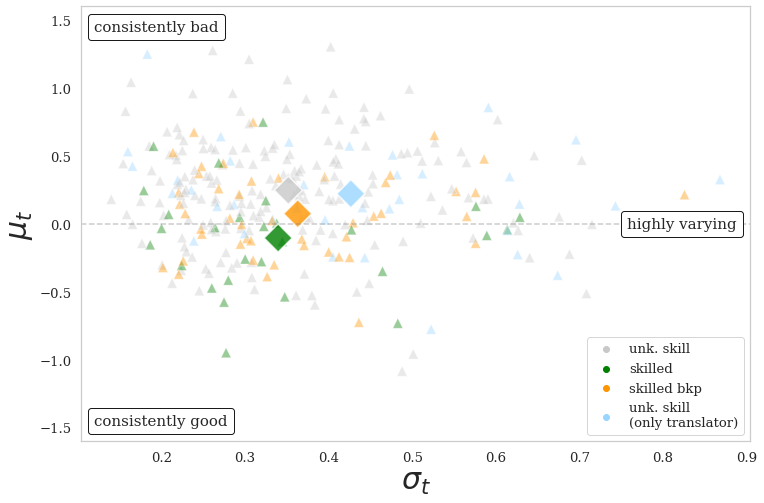}
    \caption{Translators: the lower $\sigma_T$, the more consistent the effect of the translator. The lower the $\mu_T$ the higher the probability of making errors, when the translation is not perfect.}
    \label{subfig:translators_features_muT_sigmaT}
    \end{subfigure}
    \caption[Translators assessment - skills retrieval]{Translator assessment: every triangle corresponds to a translator with the colour indicating their certified skill level: green for L1 and orange for L2, while the not-skilled are split between those that act as  both reviewers and translators (grey) and those that are translators only (light blue). The squares are the centroids of the respective populations. The parameters inferred by the Hurdle model reasonably reflect external information about translators' skills. The model can reasonably retrieve translators' skills, assessing the L1 as the translators who produce fewer errors and more perfect translations - lower right side of (\subref{subfig:translators_features_muT_piT}) - and are also the most consistent - lower left side of (\subref{subfig:translators_features_muT_sigmaT}).}
    \label{fig:translators_features}
\end{figure}

We present the results of the analysis with two kinds of plots, both showing the parameters for the 4 groups described above. In Figures \ref{fig:translators_features_ci} and \ref{fig:reviewers_features_ci}, we show the 95\% CI%
\footnote{All the CI have been calculated with the bootstrap technique.}%
of the parameters for the translator and reviewer groups respectively, while Figures \ref{fig:translators_features} and \ref{fig:reviewers_features} show scatter plots comparing pairs of parameters, which enables us to characterise both translators and reviewers, respectively. For comparison, we also present the CI of the raw EPT data that has been averaged first over the translators and then over the sample of such averages. The results are first examined separately for the translator and reviewer roles, and, finally, the characteristics of those acting in both roles are presented.

\paragraph{Translators} In our Hurdle model (used here for the inference of the skill characteristics), the contribution made by the translator to the final quality EPT score of a translation job is represented by the parameter we call the translator error propensity $\epsilon_t$:

\begin{equation}
    \epsilon_t \sim \mathcal{HL}(\pi_t, \mu_t, \sigma_t)  
\end{equation}

With this, we can give the following interpretation to the parameters that rule such distribution of $\epsilon$:
\begin{itemize}
    \item $\pi_t$: the translator's contribution to the probability that a translated job will be a perfect translation ($\text{EPT}>0$). The higher the $\pi_t$, the more skilled the translator.
    \item $\mu_t$: the mean of the Gaussian distribution of $log (\epsilon_t)$. The lower the $\mu_t$, the higher we assume the translator's skill to be, reducing the probability of obtaining a high EPT value in the TQA process.
    \item $\sigma_t$: the standard deviation of the Gaussian distribution described above. The lower the $\sigma_t$, the more consistent the translator is in his error propensity.
\end{itemize}

A good translator should, therefore, have lower values of $\mu_t$ and $\sigma_t$ and higher values of $\pi_t$ when compared to the average translator.

In Figure \ref{fig:translators_features_ci}, we grouped the translators according to the populations described above and calculated the mean value of the parameters of interest, together with its 95\% CI as found by bootstrapping the sub-population of translators. 

The parameter $\mu_t$ (\ref{subfig:translators_features_ci_muT}) confirms the intuition that the most skilled translators -  L1 (green)  - should have the lowest values: it is more likely that the final translation will have a higher quality (low value of EPT) when translated by them (even excluding the separate process that governs the 0 EPT translations). Using  $\mu_t$ as the separating criterion, we can distinguish between groups corresponding to L1 (green) and L2 (orange) translators, and considerably more so between L1 and the  unknown skill groups (grey and light blue colours). This separation corresponds to the group averages of the raw EPT (c.f. Figure \ref{subfig:translators_features_ci_EPT}), despite those incorporating the joint contributions of job, translator, and reviewer characteristics.

Examining the  values of $\pi_t$ inferred (Figure \ref{subfig:translators_features_muT_piT}) we observe that the L1 are well separated from the ones with unknown skill. This, however, does not hold for L2, which leads us to suspect that some of the backups to the Level 1 translators might not have the same level of expertise.

Figure \ref{subfig:translators_features_muT_piT} shows the location of any individual translator (triangles coloured by their certified skill assignment) in the plane $(\pi_t, \mu_t)$. We display $\pi_t$ in log scale for the sake of easier visualisation. The larger squares are the mean values for the respective population (i.e., their centroids) and the dashed horizontal line separates the translators with $\mu_t<0$ (good) from the ones with positive $\mu_t$ (bad). While the orange triangles (L2) are uniformly interspersed within the bounds delimited by the unknown skill translators (grey triangles), the L1 occupy mostly the lower right area of the plane, exhibiting a high probability of producing perfect (horizontal direction) or low-error (vertical direction) translations. If we look at the population centroids (squares), we can indeed observe that the L1 group (green square) is located towards the lower-right corner of the plane, meaning that, on average, the model assigns higher values of $\pi_t$ and lower values of $\mu_t$ to the L1 group. The unknown skill groups (light blue and grey triangles) instead exhibit the opposite behaviour: most of them have higher $\mu_t$ and lower $\pi_t$ values. 

Finally, the confidence intervals on the $\sigma_t$ estimates do not allow us to state (with a 95\% confidence level) that there exists an observable difference between the populations, except in the case of the subgroup within the unknown skill grouping which corresponds to those translators that also assumed the role of reviewer (grey triangles).  This looks more consistent (lower $\sigma_t$) in its effect than the subgroup comprised of individuals who were translators only (light blue triangles). The relationship between the inferred  $\sigma_t$ and $\mu_t$ values is displayed in Figure \ref{subfig:translators_features_muT_sigmaT}: while there exists a separation between population centroids (squares) in the vertical direction, this does not hold in the horizontal direction, where the separation is minimal, excepting the translator-only subgroup (light blue).

\FloatBarrier

\section{Reviewer assessment}
Similarly to the case for translators, in the Hurdle model the effect of the reviewers is also modelled by introducing a multiplicative error into the real translation quality through a parameter that we call \textit{reviewer bias} $\beta_r$:

\begin{equation}
    \beta_r \sim \mathcal{HL}(\pi_r, \mu_r, \sigma_r)
\end{equation}

In the case of reviewers, the interpretation of the parameters that govern the distribution of $\beta_r$ is the following:
\begin{itemize}
    \item $\pi_r$: the reviewer's contribution to the probability that a translated text will be a perfect translation ($\text{EPT}=0$). Here, high values of $\pi_r$ also have a special interpretation: they correspond to the tendency of the reviewer to not pay too much attention to the review of a particular segment or to have a bias towards low severity error annotations%
    \footnote{We remind the reader that the severities \textit{Preferential} and \textit{Repetition} have zero weights in the weighted sum resulting in the final EPT score.}%
    . In other words,  they are prone to producing perfect translations, regardless of the skill of the translator involved or of the job's intrinsic difficulty.
    \item $\mu_r$: the mean of the Gaussian distribution of $log (\beta_r)$: when the reviewer finds some errors, a low value of $\mu_r$ means that they are generally \textit{lenient}, reducing the probability of the translation scoring a high EPT value in the TQA process.
    \item $\sigma_r$: the standard deviation of the Gaussian distribution of $\mu_r$. The lower the $\sigma_r$, the more consistent in their effect will the reviewer be.
\end{itemize}

\begin{figure}[!h]
    \centering
    \begin{subfigure}[t]{0.40\linewidth}
    \centering
    \includegraphics[width=\linewidth]{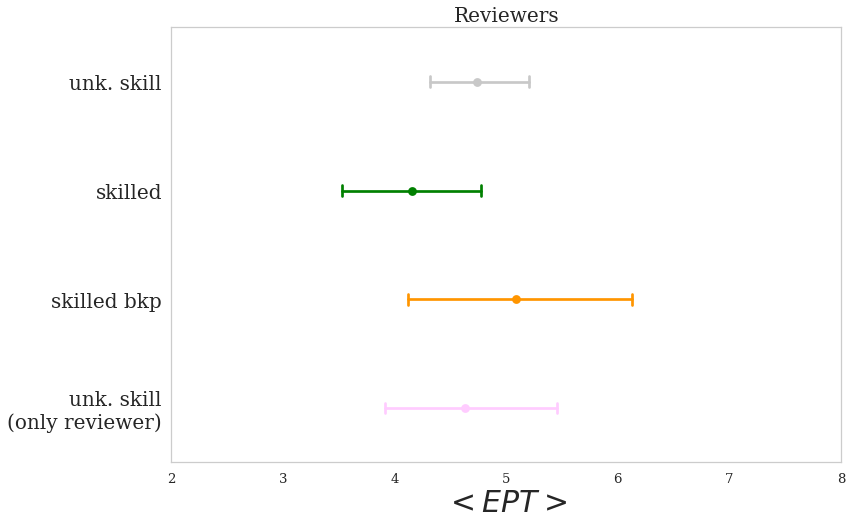}
    \caption{Reviewers: The average  of the mean EPT observed for the translations, per reviewer who generated the score.}
    \label{subfig:reviewers_features_ci_EPT}
    \end{subfigure}
    \hspace{1em}
    \begin{subfigure}[t]{0.40\linewidth}
    \centering
    \includegraphics[width=\linewidth]{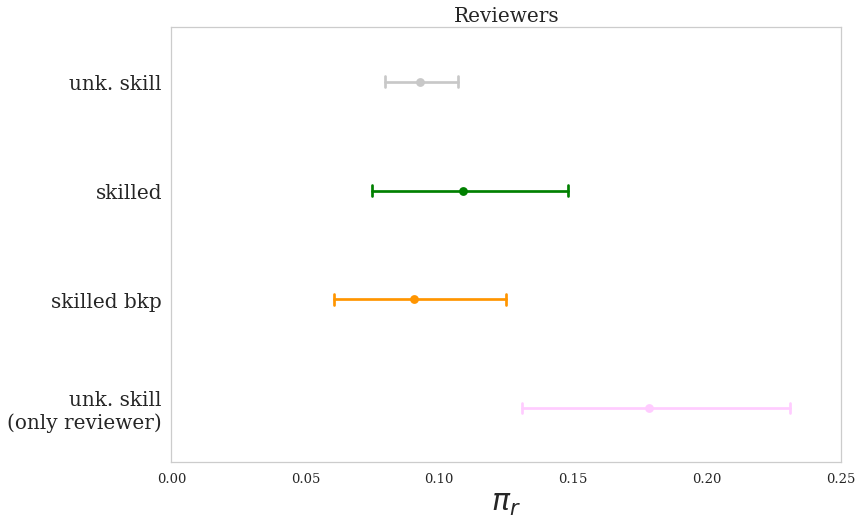}
    \caption{Reviewers: the averages of the means of the $\pi_R$ parameter per reviewer. This is a parameter of $\beta_R$, the reviewer's bias. $\pi_R$ contributes to the probability that the TQA process will produce a zero -  a perfect translation.}
    \label{subfig:reviewers_features_ci_piR}
    \end{subfigure}
    \begin{subfigure}[t]{0.40\linewidth}
    \centering
    \includegraphics[width=\linewidth]{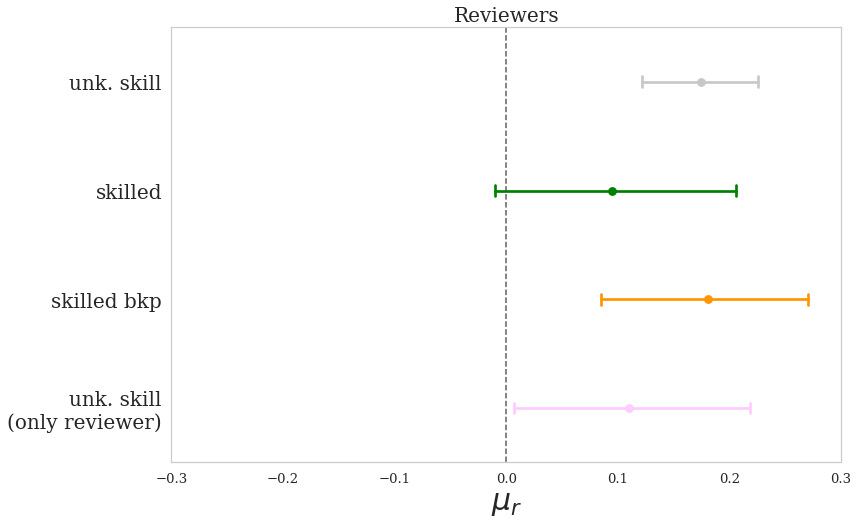}
    \caption{Reviewers: the averages of the means of the $\mu_R$ parameter per reviewer. This is a parameter of the Lognormal distribution of $\beta_R$, The higher the $\mu_R$, the higher the probability that the reviewer will annotate errors.}
    \label{subfig:reviewers_features_ci_muR}
    \end{subfigure}
    \hspace{1em}
    \begin{subfigure}[t]{0.40\linewidth}
    \centering
    \includegraphics[width=\linewidth]{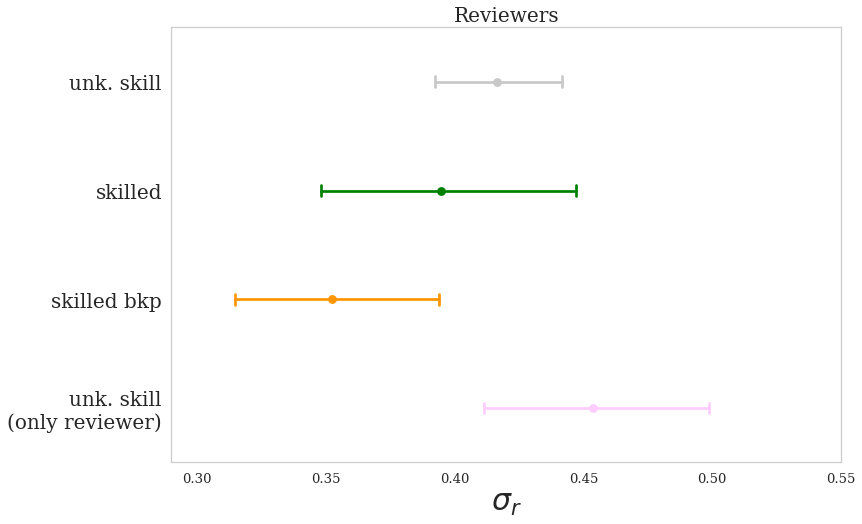}
    \caption{Reviewers:  the averages of the means of the $\sigma_R$ parameter of the Lognormal distribution of $\beta_R$. The higher value, the lower the reviewer's consistency in their effect.}
    \label{subfig:reviewers_features_sigmaR}
    \end{subfigure}
    \caption[Reviewers' assessment - CI per skill]{Reviewers assessment: 95\% confidence intervals  for the means of every parameter of the Hurdle model belonging to reviewers. The population of reviewers has been split, separating the language leads (green), their backups (orange), and the others (grey for the ones that serve both as reviewers and translators at Translated, and light blue for the ones who appear in the dataset as reviewers only.)}
    \label{fig:reviewers_features_ci}
\end{figure}

\begin{figure}
    \centering
    \begin{subfigure}[t]{.48\linewidth}
    \includegraphics[width=\linewidth]{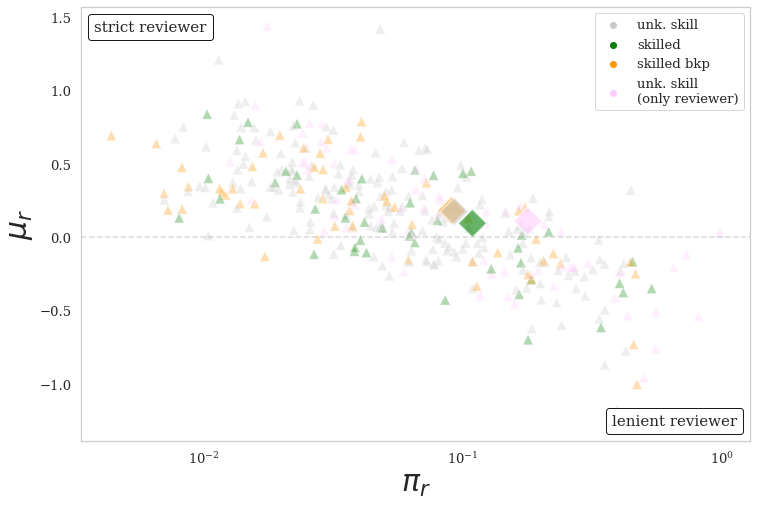}
    \caption{Reviewers: The higher the $\pi_R$, the more frequently the reviewer will evaluate a translation as perfect. The higher the $\mu_R$, the higher the chance that if the reviewer decides to annotate errors, they will find many. The pink population of those of unknown skill appears to be the least consistent in their effect, while the L2 population appears to be the strictest.
    }
    \label{subfig:reviewers_features_muR_piR}
    \end{subfigure}
    \hspace{1em}%
    \begin{subfigure}[t]{.48\linewidth}
    \includegraphics[width=\linewidth]{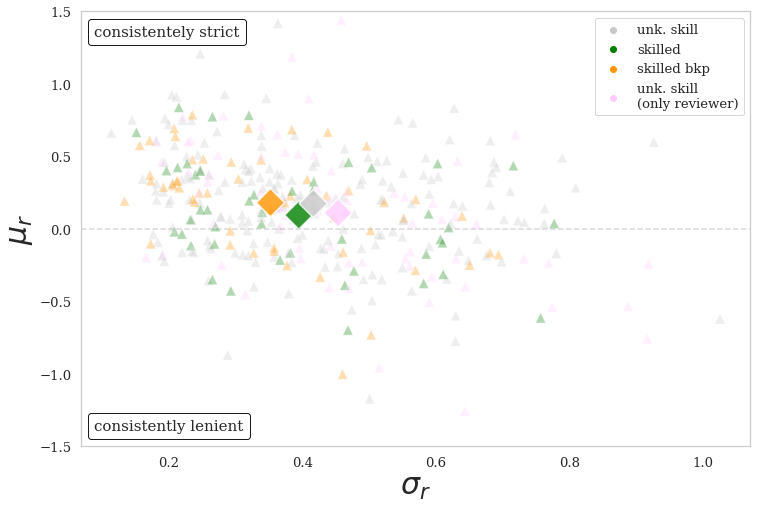}
    \caption{Reviewers: In the upper-left area of the plane lie the reviewers with a higher probability of annotating more errors (due to high $\mu_R$) while having a consistent behaviour (related to low $\sigma_R$). The L1 and L2 groups are concentrated on the left side of the plane, while in the vertical direction the L2 group and the group of linguists of unknown skill who act as both reviewers and translators seem to be the highest.}
    \label{subfig:reviewers_feaures_muR_sigmaR}
    \end{subfigure}
    \caption[Reviewers' assessment - skills retrieval]{Reviewers assessment: every triangle represents a single reviewer, and the colour indicates their certified skill level: green for L1 and orange for L2, while those of unknown skill are split between those that serve both as reviewers and translators (grey), and those that are reviewers only (pink). In the role of reviewers, the difference in the population centroids of the inferred parameters is not so significant: we can barely distinguish the L2 as stricter than the L1, for which we do not have a reasonable explanation.  Looking at $\sigma_r$,  the populations of skilled and skilled bkp seem to be more consistent in their effect than the others, especially when comparing the L2 to those of unknown skill that are reviewers only.}
    \label{fig:reviewers_features}
\end{figure}

As in the case with translators, we also show both the 95\% CI of the means of the parameter estimates (Figure \ref{fig:reviewers_features_ci}) and the location of any individual reviewer in the planes $(\pi_r, \mu_r)$  and $(\sigma_r, \mu_r)$ (Figure \ref{fig:reviewers_features}).

While the external certification  of the skills of the L1 group seems to be reflected in the model estimates of their translation skills, this doesn't translate to the case when they serve as reviewers. We cannot find a clear separation between the certified experts (L1 and L2) and the rest, at least so far as our interpretation of the parameters would suggest. This can be seen also in the average mean EPT shown in Figure \ref{subfig:reviewers_features_ci_EPT}: the skilled linguists (the certification pertains to their translation skills and not necessarily their skill as reviewers) do not exhibit a significant difference in terms of average  $\mu_r$ (Figure \ref{subfig:reviewers_features_ci_muR}) when compared to those of unknown skill. 

The mean of parameter $\pi_r$ (Figure \ref{subfig:reviewers_features_ci_piR}) seems to be higher for the group of  reviewers of unknown skill that don't act as translators in our dataset (coloured in pink). This population is composed mainly of linguists who are labelled by Translated as "preferentially reviewer"; the automatic system in charge of assigning different roles to linguists working on a translation job will subsequently prefer such linguists  when selecting a reviewer for the job. An aspect to keep in mind is that this label is assigned considering joint feedback of Translated employees and the L1 linguists, therefore such assignments could be prone to human bias: Translated employees 
could be influenced by the number of times they've observed the reviewer evaluating a translation as perfect
and could, therefore, be influenced by this in assigning the label "preferentially reviewer" to such "friendly reviewers". Such selection bias is naturally reflected in higher values of $\pi_r$ for this subgroup. This could also be the reason why the reviewers-only (pink) group in Figure \ref{subfig:reviewers_features_ci_piR} is separated from the other reviewers of unknown skill that also act in the role of translators (grey triangles). This uncertainty in the groups' separation in terms of $\pi_r$ is also fairly evident in Figure \ref{subfig:reviewers_features_muR_piR} where their means are not clearly differentiated, except for unknown skill only-reviewers (pink) along the horizontal $\pi_r$ axis.

Nonetheless, the L1 and L2 reviewers (coloured green and orange) lie mostly on the left side of the plane $(\sigma_r, \mu_r)$ (Figure \ref{subfig:reviewers_feaures_muR_sigmaR}), meaning that they are more consistent (i.e. have lower variance) than the uncertified linguists when reviewing. Although we do not expect to observe a significant separation in terms of $\pi_r$ and $\mu_r$ between the populations of reviewers examined (we have no reason to suspect a priori that a good translator is a strict or lenient reviewer), we think their values of $\sigma_r$ should reflect somehow the expertise of a linguist (even if his expertise is certified just for the role of translator): our guess is that a linguist will stabilise their style as his expertise increases, and will be less prone to inconsistency in reviewing the translations produced by others.
\begin{figure}[!h]
    \centering
    \begin{subfigure}[t]{.40\linewidth}
    \centering
    \includegraphics[width=\linewidth]{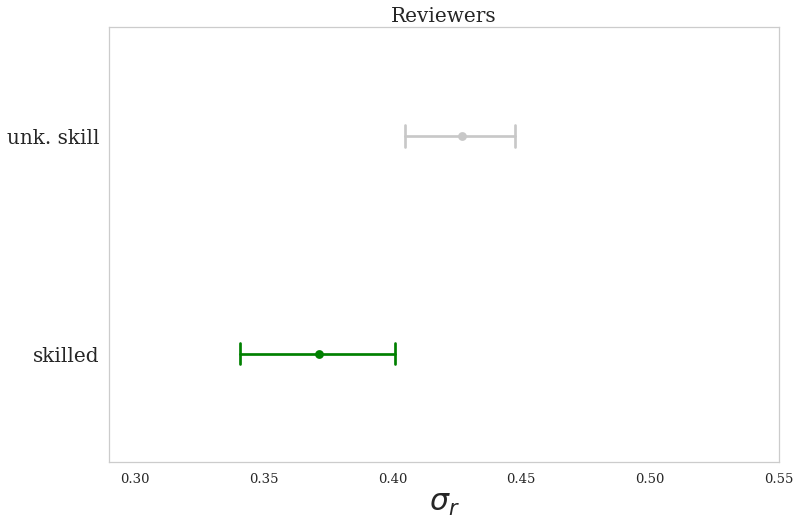}
    \caption{95\% CI of the $\sigma_r$ parameter for the two aggregated groups: skilled (L1) and skilled bkp (L2) in green and those of unknown skill and unknown skill reviewer-only in grey. If we aggregate the populations, the difference in consistency becomes significant, indicating the skilled reviewers as the most consistent.}
    \label{subfig:reviewers_features_ci_aggregated}
    \end{subfigure}
    \hspace{1em}%
    \begin{subfigure}[t]{.40\linewidth}
    \centering
    \includegraphics[width=\linewidth]{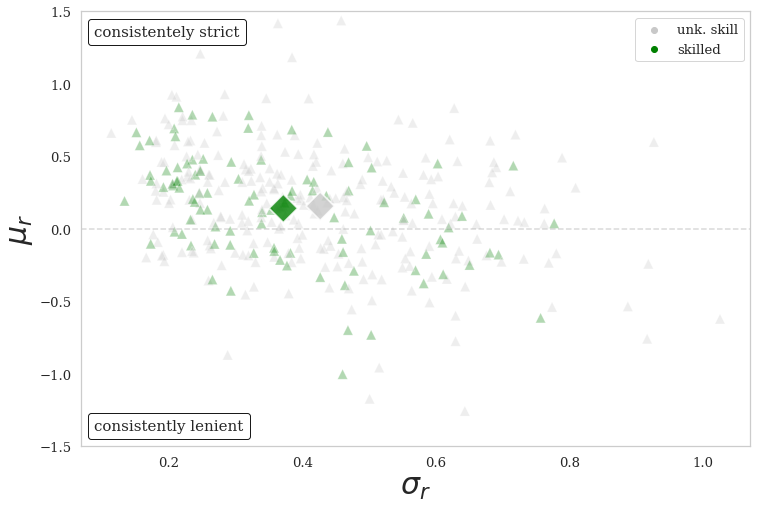}
    \caption{Same plot as in Figure \ref{subfig:reviewers_feaures_muR_sigmaR}, but with L1 and L2 grouped together, as well as unknown skill and unknown skill reviewer-only. It is particularly evident that most of the skilled linguists (green triangles) lie on the left side of the plane (i.e., they are the most consistent).}
    \label{subfig:reviewers_features_aggregated_muR_sigmaR}
    \end{subfigure}
    \caption[Reviewers assessment - skills retrieval (aggregated)]{Reviewers assessment - skills retrieval: aggregating the skilled groups (L1 and L2) and those of  unknown skill (reviewer-only and acting in both roles), the separation between the two groups becomes evident.}
    \label{fig:reviewers_features_aggregated}
\end{figure}

\begin{figure}[h!]
    \centering
    \begin{subfigure}[t]{.48\linewidth}
    \centering
    \includegraphics[width=\linewidth]{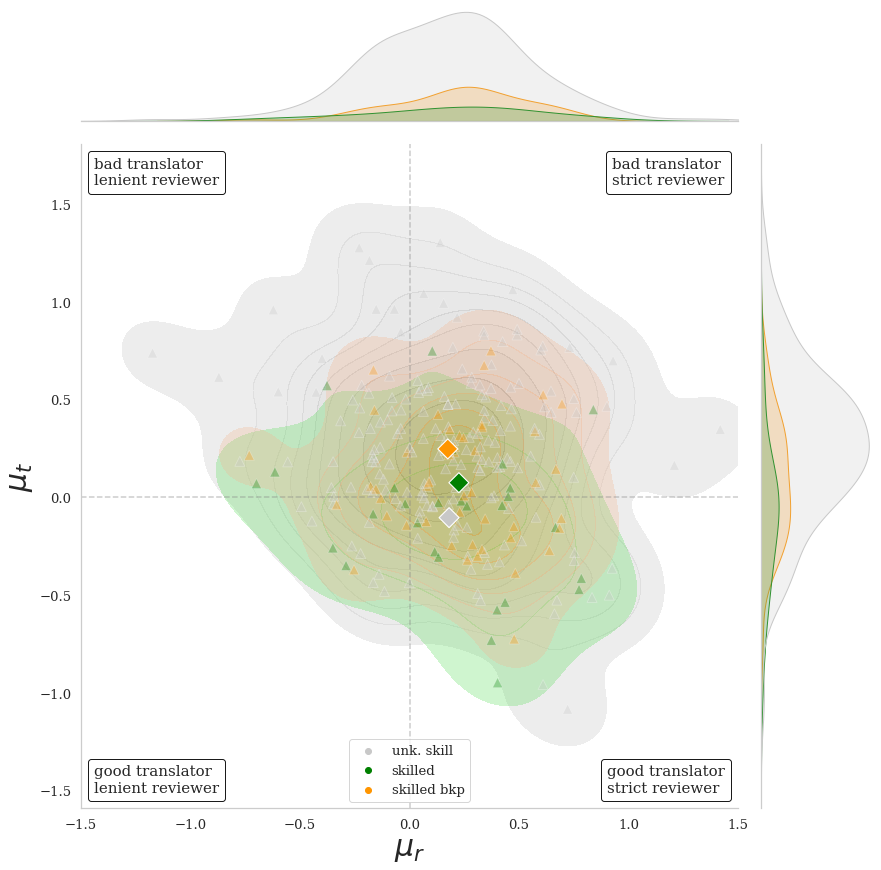}
    \caption{Linguists serving both as reviewers (horizontal direction) and translators (vertical direction) - average effect: examining both the coordinates of the group means (squares) and the areas covered by the KDE of their distributions,  we can see that while the different groups show a significant separation in terms of $\mu_t$ (vertically), they are less clearly separated in their $\mu_r$ values (horizontally).}
    \label{subfig:cross_features_mu}
    \end{subfigure}
    \hfill
    \begin{subfigure}[t]{.48\linewidth}
    \includegraphics[width=\linewidth]{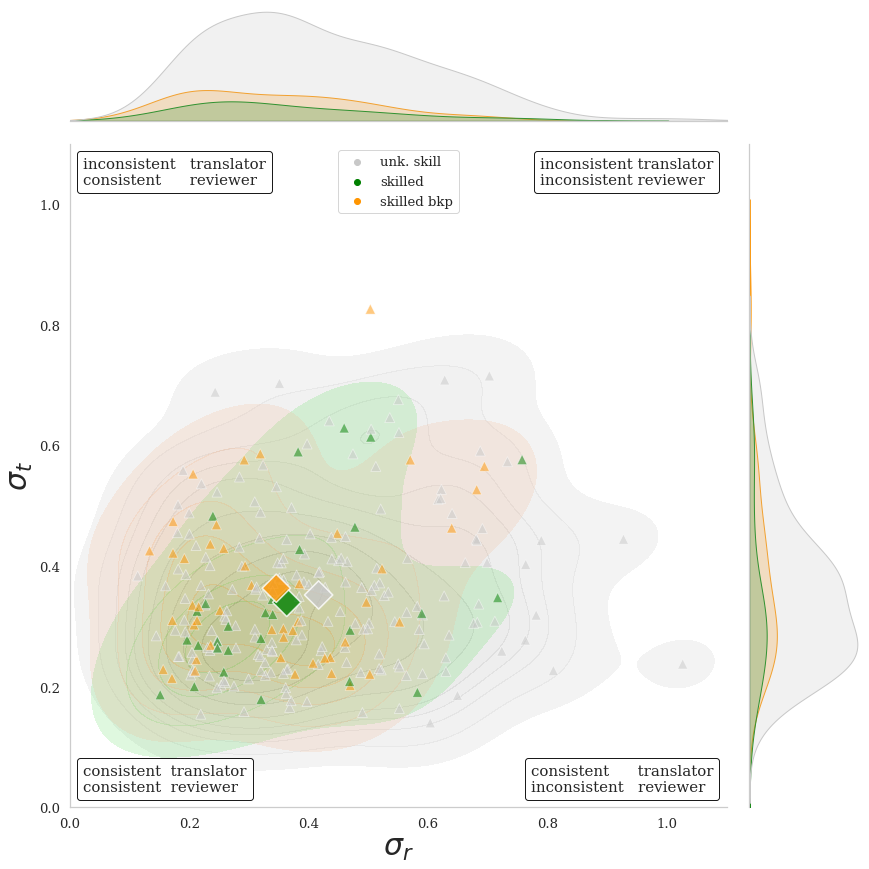}
    \caption{Linguists serving both as reviewers (horizontal direction) and translators (vertical direction) - effect consistency: the L1 and L2 groups cluster in the lower-left area of the plane, as shown by the mean coordinates of the groups, represented by the squares. Those of unknown skill are dispersed more widely in the plane.}
    \label{subfig:cross_features_sigma}
    \end{subfigure}
    \caption[Linguists assessment - cross-evaluation as reviewers and translators]{Linguists assessment - cross-evaluation as reviewers and translators: every triangle is a linguist. The colour indicates their certified skill and the squares are the mean coordinates of the linguists' groupings. The coloured area is the two-dimensional KDE of the population. The horizontal direction gives insights into the linguist's quality-assessment (i.e., review) behaviour, while  the vertical one is dedicated to their translating skills.}
    \label{fig:cross_features}
\end{figure}

We noticed that the reviewer-only category and that of unknown skill show a higher average $\sigma_r$ value compared to the average of the combined group of skilled and skilled bkp (as can be seen from the group means represented by the squares in Figure \ref{subfig:reviewers_feaures_muR_sigmaR}). We thus compared  simultaneously the CI intervals shown in Figure \ref{subfig:reviewers_features_sigmaR}, and the scatter plot in Figure \ref{subfig:reviewers_feaures_muR_sigmaR}, this time aggregating into two main groups: skilled (L1 - green - and L2 - orange - together) vs. those of unknown skill (grey and pink triangles). The result shown in Figure \ref{fig:reviewers_features_aggregated} confirms our initial intuition by inferring that the population of skilled reviewers is more consistent at a statistically significant level (lower $\sigma_r$) -  with a 95\% CI of $CI^{\text{sk}}=[0.34, 0.40]$ - than that of unknown skill, for which the CI is $CI^{\overline{\text{sk}}}=[0.41, 0.45]$.
\bigskip
\FloatBarrier
\paragraph{Cross-features} Finally, we examine the pairs of parameters $(\mu_r, \mu_t)$ for the subgroup of linguists who act as both translators and reviewers in our dataset,  aggregating L1 and L2 into one group and those of unknown skill into another  (green, orange, and grey triangles in Figure \ref{fig:cross_features}, respectively). We notice that the L1 group members cluster on the lower-left side of the $(\sigma_r, \sigma_t)$ plane (Figure \ref{subfig:cross_features_sigma}), showing a higher consistency both as reviewers and as translators when compared with those of unknown skill. Moreover, looking at the Figure \ref{subfig:cross_features_mu}, we can clearly see that while the populations of linguists are separated well as translators (in the vertical direction), we cannot say the same for their separation as reviewers (the horizontal direction), meaning that the  skill certification is reflected in the behaviour as a translator, but not necessarily so when serving as a reviewer. Therefore, we can claim that assuming that "good translator" implies "good reviewer" is a dangerous assumption to make. 
Moreover, examining the same plot (Fig. \ref{subfig:cross_features_mu}), we notice that the lower-left part of the plane $(\mu_r , \mu_t)$ is quite empty when compared to the other quadrants, which would imply that if a linguist is very good at translating (located in the lower part of the plane) they are also likely to be a strict reviewer (located on the right), and we can reasonably say that we can reasonably say this relationship holds in both directions.

Given that the groups are not strongly separated on either of the plots in  Figure \ref{fig:cross_features}, we can find a considerable portion of linguists, who, despite not being certified by Translated as skilled, actually exhibit the behaviour of a skilled linguist as indicated by the values of their model parameters. It would, therefore, be of interest to check their performance case by case and to, if the findings hold up,  use the model as a quantitative metric in choosing whom to certify as a skilled linguist. The model could also help in adjusting the number of translations that Translated assigns to such linguists. When used for this purpose, the model could help improve the overall quality of delivered translations by maximising the number of translations provided by high-quality translators.

\FloatBarrier

\section{Conclusions and future work}
Translation Quality Assessment is the evaluation of the overall quality of a translation and can be influenced by many factors: the source texts do not all share the same difficulty and, even for the same text, the difficulty can vary with the choice of the target language into which the text is translated.
The final translation quality will reflect the skill of the translator who produced it, and its assessment will be further affected by the characteristics of the reviewer, who can be strict or lenient. Both the translator and the reviewer can also have a higher or a lower level of consistency: the translator in the number of errors they produce, and the reviewer in their own judgements. 

Our work aims to face the lack of reliability of the TQA process  when conducted by humans due to human subjectivity affecting both the evaluation and the translation itself. 
We have taken into account three main features contributing to the assessment of a translation's quality, namely the job's difficulty, the translator's skills, and the reviewer's characteristics.  We treated the observed data about the quality as one measurement of a real latent quality score so as to consider the contribution of a translation job's intrinsic difficulty as well as that of one of the two human linguists involved in the evaluation process as errors affecting such measurements.
In order to obtain insights into the TQA process, we parameterised the above errors introduced into the quality assessment by creating two probabilistic models, differing in their choice of the parameters and in their likelihood functions that links such parameters to real data.
While the simplest Gaussian model (similar to the one used by \cite{Mathur2018TowardsAnnotators}) does not consider the effect of the glut of perfect (EPT=0) translations apparent from the data observed, the more complex Hurdle model does. 
Furthermore, the Hurdle model considers the errors in the measurements of the quality-score as multiplicative, while the Gaussian model treats such errors as additive. The better fit to empirical data of the Hurdle model suggests that the multiplicative relation is a better choice for modelling the systematic error in the TQA evaluations. 
Our models are shown to be consistent with existing external knowledge about translator skill, and additionally, provide useful insights into the behaviour of the reviewers who work on the TQA process. 
In particular, the results obtained from the Hurdle model indicate that skilled translators are usually stricter when evaluating a translation job as reviewers, implying that the increasing expertise in translation could induce a bias when assessing translations produced by others, even if such expertise ensures a certain level of assessment consistency.
For this reason, we claim that it is not safe to rely on a reviewer's performance to be guaranteed by their translation expertise. It is necessary, instead, to evaluate their characteristics in the role of a quality assessor alone, as we did  by modelling reviewers' bias and consistency.
Finally, our Bayesian approach is shown to perform well despite often having just one review per translation job in the data to work with: it can, nevertheless, retrieve plausible patterns of the TQA process and assess the characteristics of the translators and reviewers involved. This contrasts with approaches requiring several reviews per translation, which are more expensive both in terms of time and money. 

The current bottleneck in our method is the single translation job quality measurement we receive from the data: while the models can see and evaluate the performance of the same reviewers and translators across multiple jobs, they have only one piece of information about a specific translation job. The remaining uncertainty in translation quality is still too high to catch the fluctuations between different texts. For future improvements in this estimation, we could consider collecting more than one review per job (creating new data specifically for the purpose of calibrating models) or we could use the properties of the text itself in order to build a more informative prior for a translation's intrinsic difficulty. 
As an example of the latter approach, we could measure the "entropy of translation" \citep{zhao2019entropyUnder-translation, dimitrova2005, Hale2006Uncertainty, Krings2001, MMTeich2017EntropySurprisal}, i.e. the degree of unpredictability of the translation outcome given the source text, in order to have a more suitable prior assessment of a specific translation's intrinsic difficulty, which would decrease the uncertainty of the model about its quality assessment outcome. 

Finally, our modelling is affected by finite-size effects since we use the EPT metric, a weighted ratio of counts. An extension of this work would use both error and word count instead of the EPT; we could also model the documents at the segment level instead of as a whole.
\newpage

\section*{Acknowledgements}

This project was funded by Translated SRL, a leading provider of professional translation services. MM and SB are grateful to Marco Trombetti, CEO of Translated, for continuous support and feedback, especially on validation, as well as to several Translated colleagues: to the Core product team for help with data acquisition and curation; to the Quality Management team with validation and mapping to business needs; to the Project Management team for discussion of results and feedback; to Marcelo Albuquerque for feedback on this work and research contributions outside the scope of this article; to Anil Keshwani and Jonas Mack for technical proofreading. All authors are indebted to Marco Trombetti and Luciano Pietronero, Head of Centro Ricerche Enrico Fermi, Rome, for instigating the institutional collaboration which made this research project possible.

\section*{Author contributions}
Conceptualization: SB; Methodology, formal analysis: all; Investigation: MM, AL; Data curation: MM, AL; Software: MM, AL; Validation: MM, AL; Writing original draft, visualisation: MM; Review and editing: all; Supervision: SB, AZ.
\bibliography{references}

\end{document}